\PassOptionsToPackage{pagebackref}{hyperref}
\documentclass[10pt,twocolumn,letterpaper]{article}
\usepackage[pagenumbers]{cvpr}
\usepackage{graphicx,float,amsmath,amssymb}
\usepackage{tikz}
\usepackage{algorithm}
\usepackage{algorithmic}
\usepackage{soul}
\usepackage{color} 
\usetikzlibrary{arrows.meta}
\usepackage{caption}
\usepackage{subcaption}

\usepackage{pgfplots}
\usepgfplotslibrary{groupplots}
\pgfplotsset{compat=1.18,
  moodaxis/.style={
    font=\small,
    tick label style={font=\footnotesize},
    label style={font=\small},
    title style={font=\small},
    axis lines=left,
    axis line style={line width=0.4pt},
    tick style={line width=0.4pt},
    tick align=outside,
    major tick length=2pt,
    scaled ticks=false,
    grid=none,
    legend style={draw=none,font=\footnotesize,fill=none},
    legend cell align=left
  }
}

\definecolor{cvprblue}{rgb}{0.21,0.49,0.74}
\usepackage{hyperref}
\hypersetup{breaklinks,colorlinks,allcolors=cvprblue}
\hypersetup{pdftitle={LunarGloc: Direct Lunar Global Localization with Renderer in the Loop},pdfauthor={Chee-Kheng Chng, Sarah Dickinson, Ono Tatsuharu, Marcus Märtens, Yasuko Kasai, YUKIHIRO TAKAHASHI, Tat-Jun Chin}}
\def\confName{CVPR}
\def\confYear{2027}
\title{LunarGloc: Direct Lunar Global Localization with Renderer in the Loop}
\author{Chee-Kheng Chng$^{1,*}$\quad Sarah Dickinson$^{1}$\quad Ono Tatsuharu$^{2}$\quad Marcus Märtens$^{3}$\\
Yasuko Kasai$^{3}$\quad Yukihiro Takahashi$^{2}$\quad Tat-Jun Chin$^{1}$\\[3pt]
{\normalsize $^{1}$\href{https://adelaide.edu.au}{Australian Institute for Machine Learning, Adelaide University}}\\
{\normalsize $^{2}$\href{https://hokudai.ac.jp}{Hokkaido University}\qquad $^{3}$\href{https://isct.ac.jp}{Institute of Science Tokyo}}\\[1pt]
{\small $^{*}$Corresponding author: \href{mailto:chngcheekheng@gmail.com}{\texttt{chngcheekheng@gmail.com}}}}
\begin{document}
\maketitle

\begin{abstract}
Remote sensing of the Moon's surface from orbit is vital to further our understanding of its composition and support future exploration. Advanced remote sensors however require highly accurate positioning of the Moon-orbiting spacecraft that radar tracking from Earth is unable provide due to the distances involved. Vision-based navigation has emerged as a compelling approach to achieve lunar global localization. However, existing approaches rely on detecting and matching natural landmarks on the surface such as craters, which are impractical over landmark-poor regions and brittle due to the need to accurately detect and segment the observed landmarks. In this paper, we propose a direct lunar global localization method that employs all pixel information in the query image to determine the pose of the spacecraft. Central to our method is using a physically accurate lunar rendering engine and a pretrained foundation model to overcome the sim-to-real gap to achieve robust matching across diverse lunar terrains. An efficient Monte Carlo algorithm that conducts global search with the renderer in the loop is developed to precisely localize the query image. Results on a novel real lunar image dataset indicates the accuracy and computational efficiency of our method.
\end{abstract}

\section{Introduction}
\label{sec:intro}

\label{sec:intro-task}

Interest in lunar exploration rapidly growing, and the major space agencies have concrete ambitions~\cite{NASAArtemis,CNSAChangeProgramme,ISROChandrayaan4,JAXALUPEX} to return to the Moon in the next decade. Planned activities include investigating under-explored regions (\eg, lava tubes, permanently shadowed craters), \emph{in situ} resource utilization, and establishing long-term human presence. Underpinning the activities is estimating a global map of the surface composition via remote sensing orbiters~\cite{Chin2007LRO,Pieters2009M3,Kobayashi2025TSUKIMI}.

Critical to the quality of orbital mapping is the accurate global positioning of the spacecraft. More exquisite sensing devices typically demand positioning accuracies that exceed the ability of cislunar radar tracking alone~\cite{McLeod2025STELLA, Mazarico2018LROOrbit}. Vision-based terrain-relative navigation (TRN)~\cite{Johnson2008TRNSurvey}, which aims to localize a spacecraft based on an image of the Moon's surface, has emerged as promising back-up or alternative.

\begin{figure}[!t]\centering

    \begin{subfigure}[b]{0.23\textwidth}
         \centering
         \includegraphics[width=0.99\linewidth]{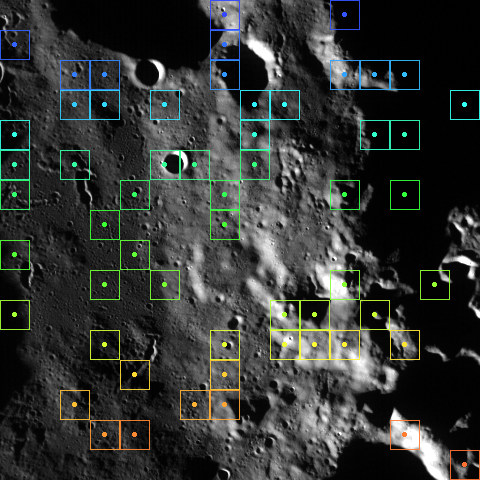}
    \end{subfigure}
    \begin{subfigure}[b]{0.23\textwidth}
         \centering
         \includegraphics[width=0.99\linewidth]{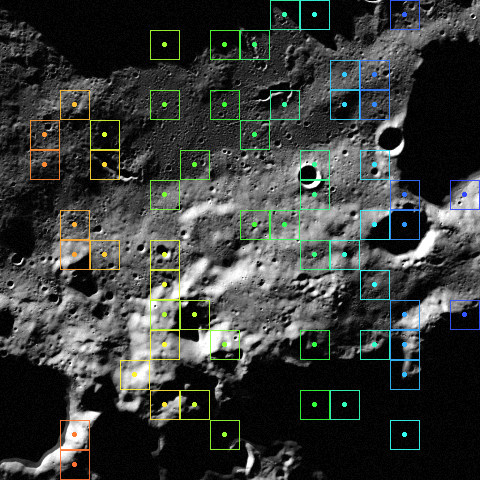}
    \end{subfigure}
    \begin{subfigure}[b]{0.23\textwidth}
         \centering
         \includegraphics[width=0.99\linewidth]{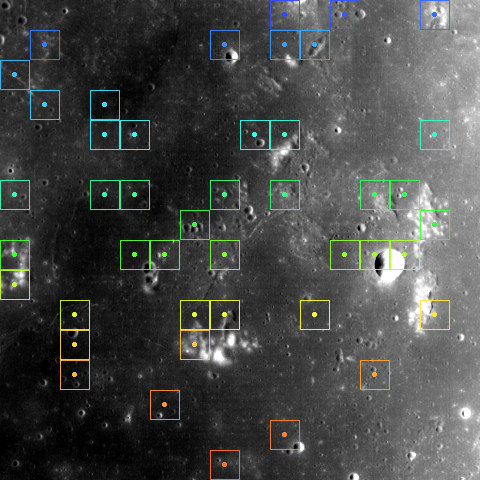}
         \caption*{(Left) Real query images.}     
    \end{subfigure} 
    \begin{subfigure}[b]{0.23\textwidth}
         \centering
         \includegraphics[width=0.99\linewidth]{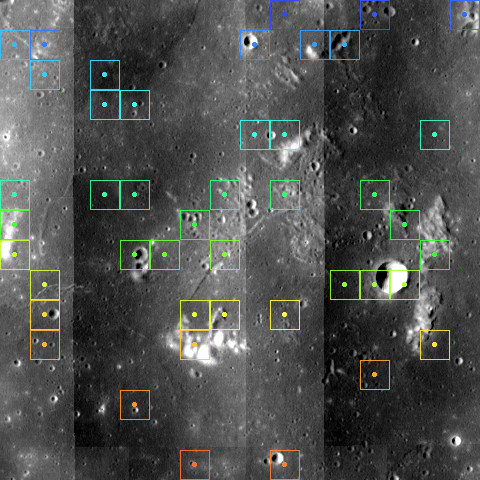}
         \caption*{(Right) Matched rendered images.}         
    \end{subfigure}
  \caption{At its core, LunarGloc conducts direct matching of real query images (left) with rendered gallery images (right). These examples correspond to texture-poor regions under difficult lighting; moreover, the rendered images show discontinuities due to efficiency-driven approximations in the rendering. Despite such challenges, we show that pretrained DINOv2 features can accurately match lunar images across differences in orientation, rendering artifacts and sim-to-real gaps. Above, patch correspondences correctly identified using DINOv2 are of the same color.}
  \label{fig:query_rendered}
\end{figure}

Crater-based navigation (CBN)~\cite{SimardBilodeau2012Pinpoint, Maass2020CraterNavigation, Christian2021CraterIdentification} is a notable example. It detects craters in the query image, matches the detections to a catalog of known craters, and estimates the camera pose from the 2D-3D crater correspondences~\cite{Maass2020CraterNavigation,McLeod2024RobustPnC,McLeod2025STELLA}. CBN is enabled by the fact that a crater rim can be modeled as a 2D ellipse and that large surveys of lunar craters are available~\cite{Robbins2019Catalogue}. However, being a \textit{feature-based} localization method, CBN will fail in regions with sparse craters. Note that crater density is irregular across the lunar surface~\cite{Zhao2023Volcanism}. CBN is also brittle, since inaccuracies in the individual steps of crater detection/segmentation and ellipse fitting will propagate through the pipeline. This fragility is amplified by challenging lighting conditions, which cause poor contrast in the query image. Fig.~\ref{fig:query_rendered} shows real lunar images that depict difficult cases for CBN.

In contrast to CBN, \emph{direct} localization uses all available measurements (intensity values) in the query image to determine the camera pose, specifically by comparing the query to a corpus of geo-annotated surface images. Besides obviating the need to detect and mathematically model specific natural landmark types, direct localization can leverage less salient cues on the terrain such as ridges, peaks and shadows. However, two basic issues affect direct methods:
\begin{itemize}
    \item There is a dearth of truly planetary scale corpora of \emph{real} geo-annotated lunar surface images with diverse conditions in altitude, lighting, viewpoints, \etc.
    \item Directly comparing lunar surface images can be difficult due to the self-similar and often texture-poor lunar terrain.
\end{itemize}
We speculate these reasons have led to inadequate attention thus far on direct methods for lunar global localization.

\vspace{-1em}
\paragraph{Contributions}

We present \emph{LunarGloc}, a direct global lunar localization technique for vision-based TRN for lunar orbiters. The main characteristics of LunarGloc are:
\begin{itemize}
    \item Inspired by render-and-compare localization methods~\cite{Trivigno2024MCLoc, Wu2024UAVD4L, Maggio2023LocNeRF, Berton2024MeshVPR, Poglitsch2015ParticleFilter}, LunarGloc exploits physically accurate lunar rendering engines~\cite{Marshall2025BOROMIR,Martin2019PANGU} to overcome the lack of real image corpora. The renderer is embedded in the search algorithm to enable precise global localization.
    \item LunarGloc leverages advances in vision foundation models (VFM) for  visual place recognition (VPR)~\cite{Sarlin2019Hierarchical,Keetha2023AnyLoc, Izquierdo2024SALAD, Garg2024SegVLAD}. Specifically, we show that pretrained DINOv2 embeddings~\cite{Oquab2023DINOv2} can  overcome the sim-to-real gap between query and rendered lunar images and achieve robust comparisons across diverse lunar terrains; see Fig.~\ref{fig:query_rendered}.
\end{itemize}
Connecting the key characteristics above is a novel processing pipeline that combines image- and patch-level DINOv2 descriptors to attenuate nuisance lighting effects and enhance the global specificity of the embeddings. The pipeline also conducts descriptor-guided Monte Carlo search with rendering by projection to reduce rendering costs and deliver planetary-scale localization in a matter of minutes.

To evaluate the robustness of LunarGloc, we developed a real lunar image dataset with ground truth poses for the global localization task: \emph{LunarGloc5k}. The dataset comprises $5,000$ images collected from observations by the wide angle camera (WAC) on the Lunar Reconnaissance Orbiter (LRO) mission~\cite{Robinson2010LROC, Robinson2009LROCCDR}. As far as we are aware, a dataset of this size for lunar global localization is unique\footnote{Our dataset will be publicly released after the peer reviewing process.}. Results show that LunarGloc outperformed CBN and SOTA VPR-based large-scale localization techniques~\cite{Trivigno2024MCLoc, Wu2024UAVD4L} in terms of accuracy, runtime and rendering cost\footnote{A demo program for LunarGloc will be released if accepted.}.

\section{Related Work}
\label{sec:related}

\subsection{Crater-based navigation}
\label{sec:related-feature-lunar}
CBN has been actively researched for decades~\cite{SimardBilodeau2012Pinpoint,Maass2020CraterNavigation,Downes2020LunaNet,Christian2021CraterIdentification,Daftry2023LunarNav,McLeod2025STELLA}. These localization systems target different aspects of a lunar mission, for instance, landing~\cite{Maass2020CraterNavigation}, orbiter~\cite{McLeod2025STELLA, McCleary2025LunarVoodoo}, and rover navigation~\cite{Thomas2025LunarLoc}. 

\begin{figure}[!t]
\centering
\includegraphics[width=0.4\linewidth]{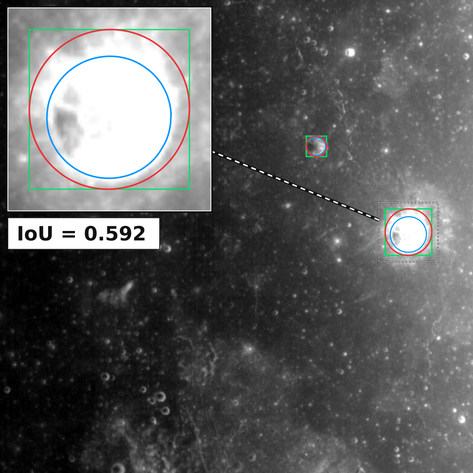}
\includegraphics[width=0.4\linewidth]{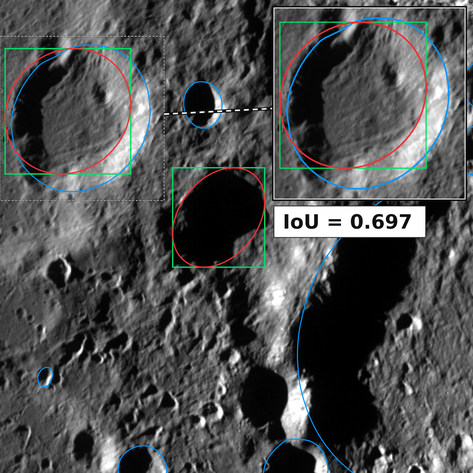}\\
\vspace{-0.5em}
\caption{Challenging illumination can obscure crater rims and make their boundaries ambiguous. Enlarged crops compare fitted ellipses (red) with ground-truth (blue), highlighting fitting uncertainty even when craters are detected (green boxes).}
\label{fig:challenging_ellipse}
\end{figure}

\vspace{-1em}
\paragraph{Front-end.} The front-end part of the pipeline detects craters and extracts their ellipse representation for the downstream tasks. Most recent crater detection modules adopt a deep learning-based approach~\cite{Downes2020LunaNet, LaGrassa2023YOLOLens, Giannakis2024SAMCrater, Zhang2025YOLOCrater,mahendrakar2026detecting}. 

A crucial problem for developing learning-based perception modules in the space domain is the lack of real training data. Until the recent effort by NASA~\cite{Fraccaro2026LunarFM}, no large-scale real lunar dataset with annotations for crater detection existed. Consequently, most methods were trained on synthetic data~\cite{McLeod2025STELLA, Zhang2025YOLOCrater, Catalan2025Artemis}. We observed a significant gap between synthetic- and real-image-trained models in our experiments (see Sec.~\ref{subsec:results}). 

Facing similar data scarcity, multiple space domain problems seek solutions in transferring the knowledge learned from large-scale Earth data. Researchers are actively investigating transfer learning for lunar terrain perception, including fine-tuning pretrained models for 3D reconstruction~\cite{Grethen2025StereoLunar} and relative surface height estimation~\cite{Bauer2026LunarHeight}, as well as distilling lunar-adapted models into smaller networks~\cite{Grethen2026LunarDistillation}. However, LuMon finds that fine-tuning on synthetic data can improve in-domain depth estimation while yielding limited transfer to real Chang'e-3 imagery~\cite{Sekmen2026LuMon}. 

These works reveal an underexplored regime: a \emph{training-free VFM-based solution}. We find that this is particularly fitting for lunar localization since the VFM is well-explored under the VPR setting in the Earth domain. 

\vspace{-1em}
\paragraph{Back-end.} There are two main modules in the back-end: crater identification (CID) and pose estimation. The crater identification task is equivalent to the VPR task in large-scale localization. It seeks the catalogd craters with the highest similarity to the detected craters. The final module then estimates the pose based on the detected-catalogd crater correspondences. 

Broadly speaking, much of the effort in this area is devoted to designing descriptors that are robust to the front-end noise~\cite{Chng2024PECAN, McLeod2024RobustPnC, Christian2021CraterIdentification}. Indeed, we observe that the back-end modules are highly sensitive to the front-end outputs, namely the ellipse fit on the detected craters. Fig.~\ref{fig:challenging_ellipse} depicts this challenge by showing the ambiguous crater-rim outlines. This discrepancy yields large localization errors, as we show in Sec.~\ref{subsec:results}. 

\subsection{Large-scale visual localization methods}
\label{subsec:large-scale}

\paragraph{Visual place localization (VPR).} Large-scale visual localization commonly starts by retrieving candidate views from a static reference database~\cite{Sarlin2019Hierarchical,
Wu2024UAVD4L,Berton2024MeshVPR}, whose coverage limits the viewpoints available for matching. AnyLoc~\cite{Keetha2023AnyLoc} aggregates pretrained DINOv2 descriptors~\cite{Oquab2023DINOv2} into a global image representation and reports lower retrieval accuracy on unstructured
terrain than on urban and indoor scenes. Patch-NetVLAD~\cite{Hausler2021PatchNetVLAD} and EffoVPR~\cite{Tzachor2025EffoVPR} use patch correspondences to distinguish candidates with similar global appearance. Motivated by these findings, we study how well image-level similarity and geometrically consistent patch matches identify the same lunar terrain in real orbital images and rendered views without lunar fine-tuning.

\vspace{-1em}
\paragraph{Static database.} A static reference database is commonly adopted in large-scale VPR-based localization~\cite{Sarlin2019Hierarchical,Wu2024UAVD4L, Berton2024MeshVPR}. This paradigm is more appropriate for re-localization tasks where the database is constructed with real images covering the scene. The success rate and accuracy of this approach depend heavily on database coverage. Nevertheless, it remains a feasible pipeline, and we compare against the latest method in this paradigm to contextualize our performance.

\vspace{-1em}
\paragraph{Online rendering.} Render-and-compare methods synthesize candidate views during
localization~\cite{Yan2023Render2Loc}. Several methods refine a pose
using a learned scene representation~\cite{YenChen2021iNeRF,
Botashev2024GSLoc,Chen2024NeFeS}, while Loc-NeRF~\cite{Maggio2023LocNeRF}
performs Monte Carlo localization in a pretrained NeRF map.
MCLoc~\cite{Trivigno2024MCLoc} is closer to our setting because it
renders directly from a supplied 3D mesh. Starting from poses retrieved
by VPR, it ranks rendered views by feature similarity to the query and
samples new poses around the best candidates. LunarGloc uses a 3D lunar
surface model~\cite{Barker2016SLDEM,Neumann2010LOLAGDR} and samples across the search region.
Inspired by EarthMatch~\cite{Berton2024EarthMatch}, it also fits a 2D
similarity transform to matched DINOv2 patches to guide the next
camera-position sample.

\subsection{Lunar Datasets}
\label{sec:related-lunar-datasets}
CRESENT and CRESENT+~\cite{McLeod2024RobustPnC,McLeod2026GeometricPrimitives}
provide synthetic orbital views for crater-based pose estimation,
while CRESENT-365~\cite{McLeod2025STELLA} adds near-global coverage
and varying illumination. Luna-1~\cite{Chase2024MARs} provides
synthetic orbital imagery for crater-landmark recognition, and
LuSNAR~\cite{Liu2024LuSNAR} provides synthetic rover observations
for localization, mapping and scene understanding.

Lunar Studio, introduced with LNEM~\cite{Lee2026LNEM}, uses real
orbital imagery for terrain reconstruction. LunarPhoto, part of
MoonAnything~\cite{Grethen2026MoonAnything}, pairs real,
orthorectified NAC imagery with renderings for photometric
perception. Rodda et al.~\cite{Rodda2024CE5Crater} annotate
Chang'e-5 descent images for crater detection.
POLAR Traverse~\cite{Hansen2024POLARTraverse} provides stereo
sequences from a physical lunar-analogue testbed under simulated
polar lighting for visual odometry and mapping.
LunarGloc5k complements these resources with geographically
distributed perspective queries constructed from real orbital
observations for single-image global localization.

\section{LunarGloc5k}
\label{sec:data}
We introduce LunarGloc5k, a benchmark dataset for monocular global localization on the Moon constructed using real observations from LRO WAC~\cite{Robinson2010LROC, Robinson2009LROCCDR}. The main characteristics of LunarGloc5k are:
\begin{itemize}
    \item \textbf{Images and annotations.} LunarGloc5k comprises $5,000$ real query images, each annotated with its camera pose, intrinsics, and metadata associated with the source. The source images were calibrated observations from the NASA Planetary Data System (PDS) LROC archive~\cite{Robinson2009LROCCDR}. We used the $643$\,nm band of the push-frame WAC~\cite{Robinson2010LROC}. Further details can be found in the supp. material.

    \item \textbf{Geographic coverage and viewing conditions.} Camera locations span the lunar surface, including both polar regions, at altitudes of $25$ to $190$\,km and local solar elevations of $0.1^\circ$ to $85.8^\circ$.
    \item \textbf{Reference imagery.} To support VPR research, we associate each query with a set of reference images (see Sec.~\ref{subsec:baseline}) covering the entire lunar surface.
\end{itemize}

\begin{figure}[t]
  \centering
  \includegraphics[width=\linewidth]{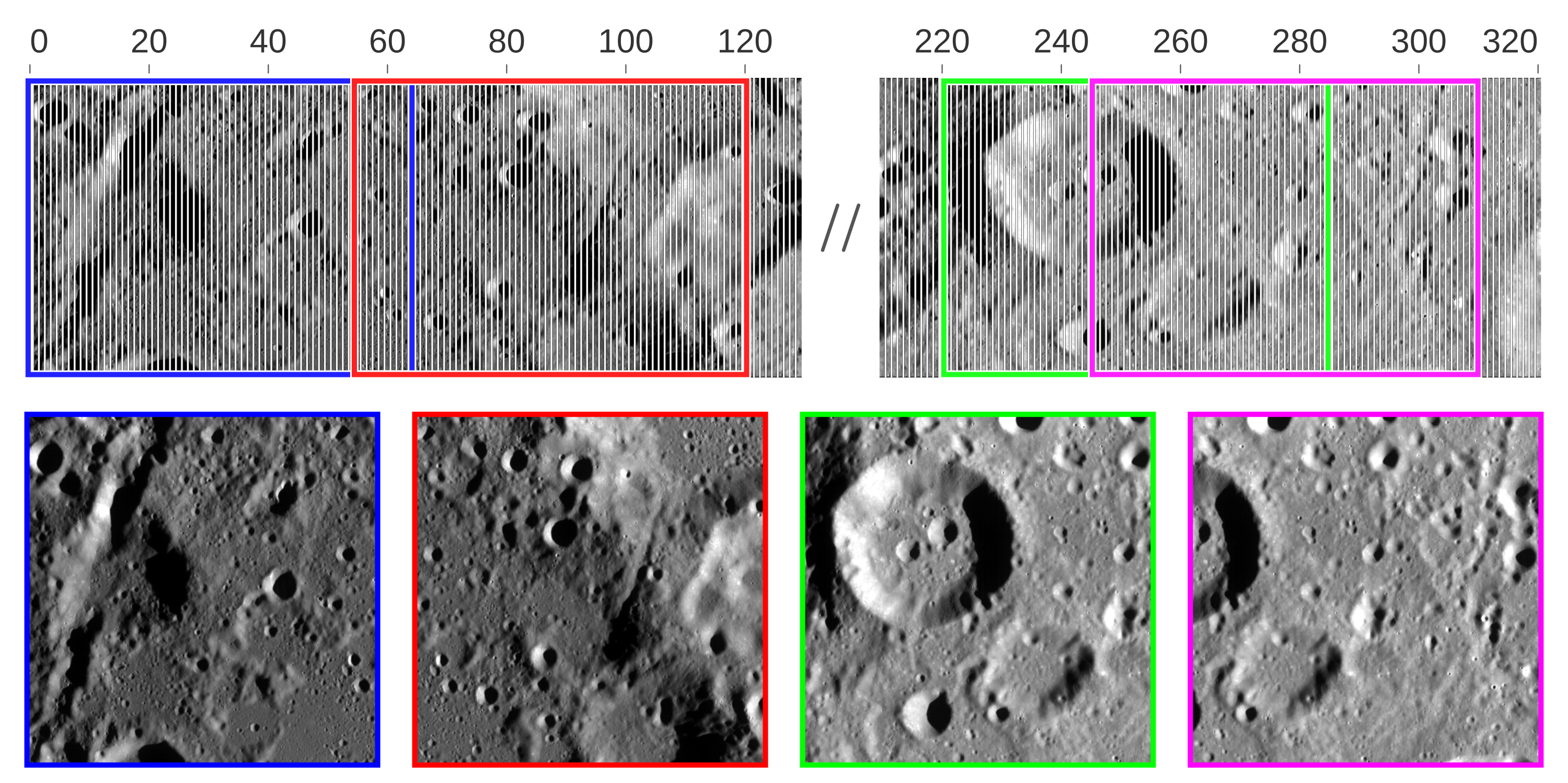}\\
\vspace{-0.5em}
  \caption{\textbf{Construction process of LunarGloc5k.} NASA LROC WAC framelets (product M171121002CC, 643 nm, 704 × 14 pixels, indexed above) and four projected query images. Colored boxes mark the source regions for the queries below. See Sec.~\ref{sec:data}.}
  \label{fig:framelet-to-query}
\end{figure}

\begin{figure}[t]
  \centering
  \includegraphics[width=0.9\linewidth]{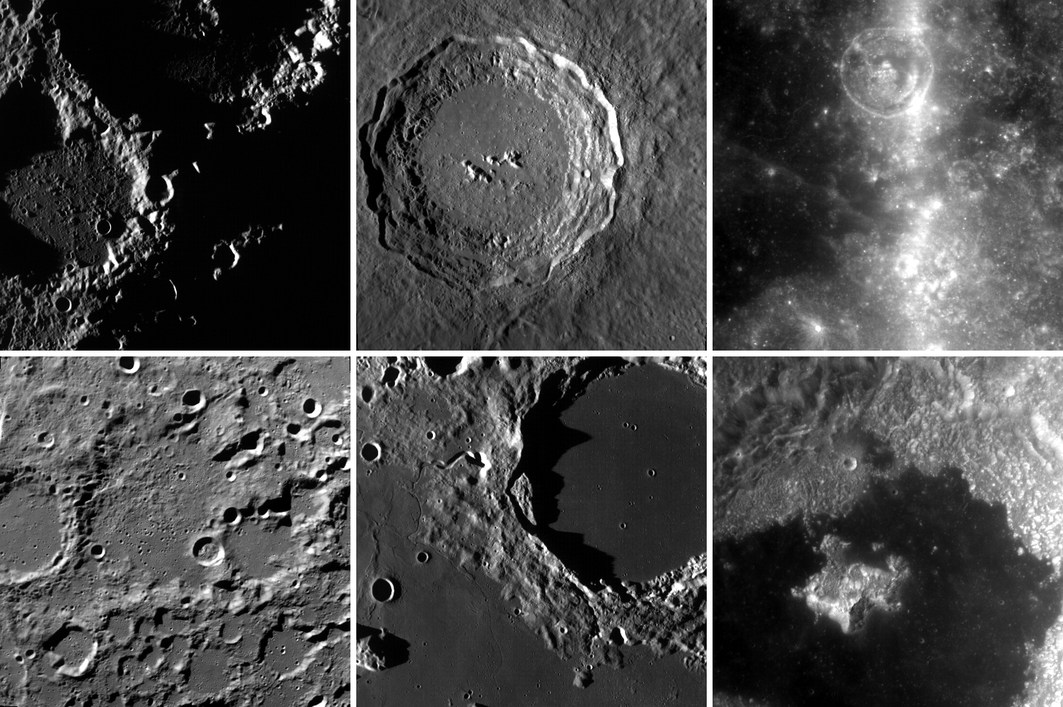}\\
\vspace{-0.5em}
  \caption{\textbf{LunarGloc5k examples} across different terrains, camera altitudes and illumination conditions.}
  \label{fig:lunarloc5k_examples}
\end{figure}

\vspace{-1em}
\paragraph{Construction}
WAC acquires neighboring framelets from different spacecraft poses.
Using the calibrated WAC camera model and the SPICE-derived pose of each
framelet~\cite{Acton1996SPICE}, we projected its pixels onto a lunar digital elevation model DEM~\cite{Barker2016SLDEM,Neumann2010LOLAGDR}
and resampled the measured intensities into a single perspective image
(Fig.~\ref{fig:framelet-to-query}).
This procedure is geometrically analogous to the camera-to-camera
reprojection performed by the ISIS \texttt{cam2cam} tool~\cite{ISISCam2Cam}.
A selected anchor framelet defines the virtual camera's center and viewing
direction. Each query has a resolution of $480\times480$ pixels and a
$60^\circ\times60^\circ$ field of view (see Fig.~\ref{fig:lunarloc5k_examples}
for examples). Further reprojection details are provided in the supp.~material.

\begin{figure}[t]
  \centering
  \includegraphics[width=\linewidth]{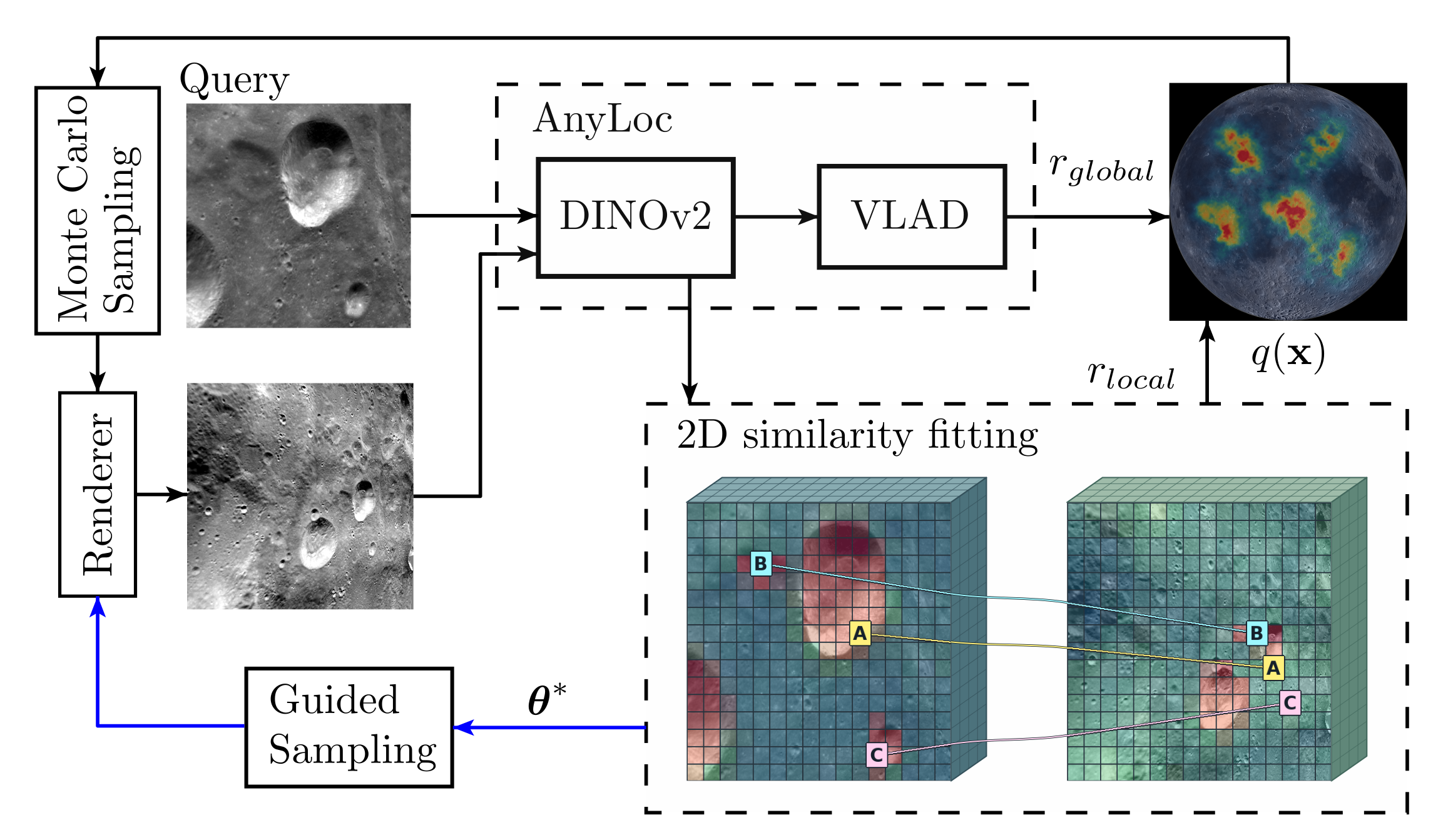}\\
\vspace{-0.5em}
\caption{\textbf{LunarGloc's global localization algorithm}. The renderer synthesizes a view at each sampled camera position $\mathbf{x}$. AnyLoc scores image-level similarity ($r_{\mathrm{global}}$), and 2D similarity fitting scores patch-level agreement ($r_{\mathrm{local}}$). Both scores update the Monte Carlo sampling density $q(\mathbf{x})$, while the fitted transform $\theta^*$ guides the next position estimate (blue loop).}\vspace{-1em}
  \label{fig:lunargloc-overview}
\end{figure}

\section{LunarGloc}\label{sec:method}

Our overall goal is finding the 6DoF camera pose in the selenographic frame~\cite{Christian2021CraterIdentification} that corresponds to the query image $I_{q}$. LunarGloc achieves this goal via two major stages: global localization and pose refinement.

\subsection{Global localization}
\label{subsec:global-localization}
The global localization stage searches for the latitude $\phi$, longitude $\lambda$ and altitude $h$ of a \emph{virtual nadir-pointing} camera that observes the similar region of the Moon as $I_q$.

Given a candidate $(\phi,\lambda,h)$, taking local north as the image-up direction, the pose of the virtual camera in the selenographic
frame is given by
\begin{equation}
\label{eq:R}
\mathbf{R}_{\phi,\lambda} =
\begin{bmatrix}
-\sin\lambda & -\sin\phi\cos\lambda & \cos\phi\cos\lambda \\
 \cos\lambda & -\sin\phi\sin\lambda & \cos\phi\sin\lambda \\
 0           &  \cos\phi            & \sin\phi
\end{bmatrix},
\end{equation}
\begin{equation}
\label{eq:t}
\mathbf{t}_{\phi,\lambda,h} = (M+h)
\begin{bmatrix}
\cos\phi\cos\lambda\\
\cos\phi\sin\lambda\\
\sin\phi
\end{bmatrix},
\end{equation}
where $M = 1737.4$\,km is the Moon's radius. The task solved by the global localization stage is thus expressed as
\begin{align}\label{eq:opt}
\begin{aligned}
& \underset{(\phi,\lambda)\in \mathcal{D}, h \in \mathcal{H}}{\text{max}}
& & f(\mathcal R(\mathbf{R}_{\phi,\lambda}, \mathbf{t}_{\phi,\lambda,h} ~ \vert~ \mathbf{C}), I_{q}),
\end{aligned}
\end{align}
where $\mathcal{D}$ and $\mathcal{H}$ denote search limits for the global coordinates and altitude, $\mathcal{R}$ renders a lunar image given 6DoF input pose $(\mathbf{R},\mathbf{t})$ and assumed priors $\mathbf{C}$ (including DEM of the lunar surface, camera intrinsics and the timestamp of $I_{q}$), and $f$ compares the rendered image with $I_q$.

Below, we describe our algorithm to efficiently solve~\eqref{eq:opt}, including the rendering and image comparison approaches. Fig.~\ref{fig:lunargloc-overview} depicts the overall pipeline of the global localization stage. Note that while~\eqref{eq:opt} solves only a 3DoF problem, we only require the rendered view to have sufficient field-of-view (FOV) overlap with $I_q$. The pose refinement stage (Sec.~\ref{subsec:pnp}) will recover the 6DoF pose of $I_q$.

\subsubsection{Monte Carlo Sampling}
\label{subsec:MC_sampling}
We sample the surface location $\mathbf{x}=(\phi,\lambda)\in\mathcal D$ adaptively and the altitude $h\in\mathcal H$ uniformly. This concentrates the search on which terrain is visible, while altitude primarily changes the scale at which it appears. After $n$ candidate evaluations, the proposal density over the lunar surface is
\begin{equation}
\label{eq:q_n}
q_n(\mathbf{x}) =
(1-\beta)\,u_{\mathcal D}(\mathbf{x}) + \beta\,g_n(\mathbf{x}),
\end{equation}
where $u_{\mathcal D}$ is uniform over the search region and $g_n$ favors locations whose rendered views resemble $I_q$. We initialize $g_0$ using the similarity scores of the pre rendered reference images, detailed in Sec.~\ref{subsec:baseline}.

To generate the next candidate, we draw $\mathbf{x}_{n+1}=(\phi_{n+1},\lambda_{n+1})\sim q_n$ and independently sample $h_{n+1}\sim\mathcal U(\mathcal H)$. These samples determine the virtual camera pose in \eqref{eq:R} and \eqref{eq:t}. Evaluating~\eqref{eq:opt} at this pose gives the similarity score $s_{n+1}$ defined in Sec.~\ref{subsec:compare}. We assign the sampled location the weight
\begin{equation}
\label{eq:candidate-weight}
a_{n+1} = \frac{s_{n+1}}{q_n(\mathbf{x}_{n+1})}.
\end{equation}
Intuitively, a location sampled from a region of higher proposal density ($q_n$) receives lower weight to prevent oversampling. We then update the adaptive density by adding a normalized spherical kernel $K(\cdot\mid\mathbf{x}_{n+1})$ centered at the sampled location,
\begin{equation}
\label{eq:guided-density-update}
g_{n+1}(\mathbf{x}) =
\frac{A_n g_n(\mathbf{x}) + a_{n+1}K(\mathbf{x}\mid\mathbf{x}_{n+1})}
     {A_n+a_{n+1}},
\end{equation}
where $A_n=\sum_{c=1}^{n}a_c$ is the accumulated weight of the existing density $g_n$, and $a_{n+1}$ determines the contribution of the new sample relative to that density.

\subsubsection{Renderer}
\label{subsec:renderer}
For each candidate $(\phi,\lambda,h)$, we render a view $I_{r}=\mathcal R(\mathbf R_{\phi,\lambda}, \mathbf t_{\phi,\lambda,h}\mid\mathbf C)$. We describe the essential details of the renderer in this section, leaving the terrain mesh construction process, which includes the source of DEM and Hapke reflectance parameters, to the supp. material. 

There are several lunar surface renderers, such as PANGU~\cite{Martin2019PANGU}, SurRender~\cite{Lebreton2021SurRender} and BOROMIR~\cite{Marshall2025BOROMIR}. We extend BOROMIR's open-source Blender Cycles renderer with the Hapke model~\cite{Hapke2012Reflectance}, which is widely used in planetary remote sensing, to improve rendering realism.  

\vspace{-1em}
\paragraph{Rendering by projection.} In practice, invoking Blender on the fly for each sampling is costly. Hence, we perform the following. First, we render a set of reference images offline to cover the search region (details in Sec.~\ref{subsec:baseline}). During localization, we `render' by projection. Specifically, we first cast a ray through each pixel of the candidate view and intersect it with the terrain represented by the DEM. Then, we project each intersection point into the pre-rendered reference images, bilinearly sample their intensities at the projected coordinates, and blend valid contributions where the reference views overlap. 

Compared to invoking Blender, this process is around 7$\times$ faster, reduced from 12s to 1.9s per render. The projected views generally preserve terrain appearance, although brightness differences between overlapping reference images can produce visible mosaic seams (Fig.~\ref{fig:query_rendered}).

\subsubsection{Compare}
\label{subsec:compare}
As alluded to in Sec.~\ref{sec:related}, we aim to explore whether a
pretrained VFM can support lunar localization without
fine-tuning on lunar imagery. However, our analysis (Sec.~\ref{subsec:in_depth}) shows that a rendered view can closely match the query in the image-level DINOv2 descriptor space even when it depicts a distant region, particularly when the two views have similar lighting. Geometrically consistent patch matches reduce this illumination bias, so we score each candidate using both global appearance and local geometric similarities:
\begin{equation}
\label{eq:candidate-score}
s = r_{global} + r_{local}.
\end{equation}
Following AnyLoc~\cite{Keetha2023AnyLoc}, we aggregate the DINOv2 patch
features of each image into a single VLAD descriptor. The cosine similarity
between the query and rendered image descriptors gives the global score
$r_{global}$.

\vspace{-1em}
\paragraph{Geometric consistency of patch matches.}
To compute $r_{\mathrm{local}}$, we measure how many patch correspondences
between $I_q$ and $I_r$ agree with a single 2D similarity transform.
\begin{itemize}
\item Let $\mathcal I_q$ and $\mathcal I_r$ index the valid patches in
the query and rendered images, respectively. Let $c_{ij}$ denote the
cosine similarity between the descriptors of query patch $i$ and
rendered patch $j$. Mutual nearest-neighbor matching gives the
correspondence set
    \begin{equation}
    \label{eq:mutual-patch-matches}
    \mathcal C =
    \left\{
    \langle i,j\rangle \in \mathcal I_q \times \mathcal I_r
    \;\middle|\;
    \begin{aligned}
    j &= \operatorname*{arg\,max}_{k\in\mathcal I_r} c_{ik} \\
    i &= \operatorname*{arg\,max}_{\ell\in\mathcal I_q} c_{\ell j}
    \end{aligned}
    \right\}.
    \end{equation}
    \item Given $\mathcal{C}$, we solve the 2D similarity fitting with the maximum consensus formulation below,
    \begin{equation}
    \label{eq:similarity-fit}
    \boldsymbol{\theta}^{*}
    =
    \operatorname*{arg\,max}_{\boldsymbol{\theta}}
    \sum_{\langle i,j\rangle\in\mathcal{C}}
    \left\lfloor
    \left\|T_{\boldsymbol{\theta}}(\mathbf{p}_i^q)-\mathbf{p}_j^r\right\|_2
    \leq 1/16
    \right\rfloor ,
    \end{equation}
    where $\mathbf p_i^q$ and $\mathbf p_j^r$ are patch centers expressed
    in normalized image coordinates. On the $16\times16$ patch grid\footnote{We resize each image to $224\times224$ pixels before extracting
DINOv2 features. With $14\times14$-pixel patches, this gives $224/14=16$
patches along each axis.},
    adjacent centers along either axis are $1/16$ apart. We therefore count
    a match as an inlier ($\lfloor.\rfloor$ returns 1) if its alignment error is at most one patch spacing. The transformation function is defined as $T_{\boldsymbol{\theta}}(\mathbf{p})=\alpha\mathbf{A}\mathbf{p}+\mathbf{c}$, where the variables in $\boldsymbol{\theta}$ are a uniform scale $\alpha \in \mathbb{R}$, a planar rotation
    $\mathbf{A} \in SO(2)$, and a 2D translation $\mathbf{c} \in \mathbb{R}^2$.
    \item Finally, we define the consensus set as
    \begin{equation}~\label{eq:consensus_set}
        \mathcal{C}^{*} =
        \left\{
        \langle i,j\rangle \in \mathcal{C}
        \,\middle|\,
        \left\|T_{\boldsymbol{\theta}^{*}}(\mathbf{p}_i^q)
        - \mathbf{p}_j^r\right\|_2 \leq 1/16
        \right\}.
    \end{equation}
    We use its normalised cardinality as the local score,
    \begin{equation}
        r_{\mathrm{local}} =
        \frac{\left|\mathcal{C}^{*}\right|}{N_q} \in [0,1],
    \end{equation}
    where $N_q=16^2=256$ is the total number of patches in the query image.
\end{itemize}
    
\subsubsection{Guided sampling by 2D similarity fitting}
\label{subsec:guided_sampling}
Intuitively, the matching region of DINOv2's local patches provides clues to the location of the query image. Its location can be approximated\footnote{The 2D similarity transform cannot fully capture the image displacement in 3D. The 3D-2D PnP would yield a more accurate transform, but it is too costly to run at every iteration. Our 2D fit is $85\times$ cheaper to run.} from the parameters obtained via solving the 2D similarity fitting problem detailed in \eqref{eq:similarity-fit}.

Specifically, we transform the coordinates of the query-image center
$\mathbf{p}_c^q=[0.5,0.5]^\top$ into the normalized coordinate system
of the candidate's rendered image $I_r$ via
\begin{equation}
\label{eq:guided-image-center}
\widehat{\mathbf{p}}_c^r
= T_{\boldsymbol{\theta}^{*}}(\mathbf{p}_c^q).
\end{equation}
Using the camera intrinsics and pose of $I_r$, we backproject
$\widehat{\mathbf{p}}_c^r$ onto the lunar DEM and use the latitude
and longitude of the intersection to set the guided candidate's
coordinates $(\phi_g,\lambda_g)$. We set its altitude to $h_g=\alpha^*h_r$, where $\alpha^*$ is the fitted scale and $h_r$ is the current candidate's altitude. Finally, the guided camera pose in the selenographic frame is given by
$\mathbf{R}_g=\mathbf{R}_{\phi_g,\lambda_g}$ and
$\mathbf{t}_g=\mathbf{t}_{\phi_g,\lambda_g,h_g}$,
following \eqref{eq:R} and \eqref{eq:t}, respectively. 

The guided search operation is invoked when a potential candidate (with more than 8 inliers after solving ~\eqref{eq:similarity-fit}) is found. The candidate sampled here is used to update the proposal density~\eqref{eq:guided-density-update}. 

\subsection{Pose refinement}\label{subsec:pnp}

Once a candidate's inlier count exceeds the stopping threshold
(Sec.~\ref{subsec:hyper}), we end the global search~\eqref{eq:opt} and match SIFT
keypoints~\cite{Lowe2004SIFT} between the query $I_q$ and its rendered
candidate view $I_r$. We backproject the matched keypoints in $I_r$
onto the lunar DEM and use the resulting 2D--3D correspondences to
estimate the query camera's 6DoF pose in the selenographic frame
with RANSAC-PnP~\cite{Lepetit2009EPnP}.

\subsection{Hyperparameters}
\label{subsec:hyper}

The main hyperparameters of LunarGloc are set as follows. The $\beta$ in~\eqref{eq:q_n} is set to 0.75. We set the altitude search interval to $\mathcal{H}=[40,160]\,\mathrm{km}$, which covers most query altitudes. The surface search region $\mathcal D$ spans $45^\circ$ in latitude and $45^\circ$ in longitude for the experiments in Sec.~\ref{sec:experiments}. LunarGloc terminates the global search and begins pose refinement once a candidate's consensus set $|\mathcal{C}^*|$ contains more than 25 patch matches~\eqref{eq:consensus_set}. Minor hyperparameters are listed in the supp.

\section{Experiments}
\label{sec:experiments}
We compare LunarGloc with CBN and VPR-based localization methods on LunarGloc5k. Ablation studies assess the contributions of online sampling and patch-based guidance.

We examine illumination bias in DINOv2 retrieval and test whether geometrically
consistent patch matches improve localization recall. We assess scalability by
expanding the search from $45^\circ \times 45^\circ$ to $180^\circ \times 180^\circ$,
effectively covering the visible lunar hemisphere.

\subsection{Metrics}
\label{subsec:Metrics}
The main metrics are the position and attitude errors and runtime. Following the common practices of large-scale localization~\cite{Trivigno2024MCLoc, Maggio2023LocNeRF, Wu2024UAVD4L}, we report the recall at different accuracy levels: \textit{Coarse}: 1000\,m, 0.5$^\circ$, \textit{Fine}: 600\,m, 0.25$^\circ$, \textit{Strict}: 300\,m, 0.1$^\circ$. For the image-based methods, we also compare the number of images required by each method. 

\subsection{Baselines and competitors}
\label{subsec:baseline}
We compared with the following state-of-the-art methods discussed earlier. We selected methods that require no training to ensure a fair comparison. We used the original authors' provided implementations of all the methods below.

\vspace{-1em}
\paragraph{Crater-based localization for lunar orbiters.}
We evaluated STELLA~\cite{McLeod2025STELLA} and STELLA-LFM, which replaces STELLA's crater detector with the NASA--IBM lunar model~\cite{NASAIBM2026CraterDetection}. Both receive a $0.1^\circ$ attitude prior and initialize position from crater identification.

\vspace{-1em}
\paragraph{Static database.}
UAVLoc~\cite{Wu2024UAVD4L} is a 6DoF aerial localization method that retrieves reference views rendered offline from a 3D scene model. We also tested LunarGloc-static, which disables online sampling and
retrieves the three highest-ranked database views. We ran pose refinement(Sec.~\ref{subsec:pnp})
separately for each view and returned the query pose estimate with the
most inliers. 

Since both UAVLoc and LunarGloc-static rely on reference views that overlap the query at a similar image scale, we evaluated both on the same databases with progressively broader altitude coverage. The 1-layer database (-1) covers the search region at 100\,km, the mean query altitude. The 3-layer database (-3) adds views at 75 and 125\,km, and the 5-layer database (-5) further adds views at 50 and 150\,km.

\vspace{-1em}
\paragraph{Online rendering.}
MCLoc~\cite{Trivigno2024MCLoc} uses the same renderer as LunarGloc (Sec.~\ref{subsec:renderer}). For a fair comparison, we also applied LunarGloc's pose refinement module (Sec.~\ref{subsec:pnp}) to MCLoc. Since it has a small search radius (around 30\,km), we provided a corresponding database at the appropriate altitude for each query image. 

LunarGloc used a fixed 50\,km reference database to initialize the proposal density~ \eqref{eq:q_n} and perform rendering-by-projection (Sec.~\ref{subsec:renderer}) during the search. The altitude sampling range is defined in Sec.~\ref{subsec:hyper}. To demonstrate the efficiency of LunarGloc's guided search, we evaluated LunarGloc-w/o-guided, which disabled deterministic guidance (Sec.~\ref{subsec:guided_sampling}) while retaining the adaptive sampling. 

\subsection{DINOv2 descriptors for lunar localization}
\label{subsec:in_depth}
\paragraph{Illumination bias.}
Similar lighting can make distant lunar sites look alike. We test whether DINOv2 preferentially retrieves such sites using the absolute difference in solar elevation as our measure of illumination similarity. Our analysis in the supp. material motivates this choice by showing that DINOv2 descriptors are more sensitive to changes in solar elevation than in solar azimuth. For each query, we rank reference views of the visible lunar surface, rendered at the query timestamp, and retain the top-5 under each score.

\begin{figure}[t]
    \centering
    \includegraphics[width=0.49\linewidth]{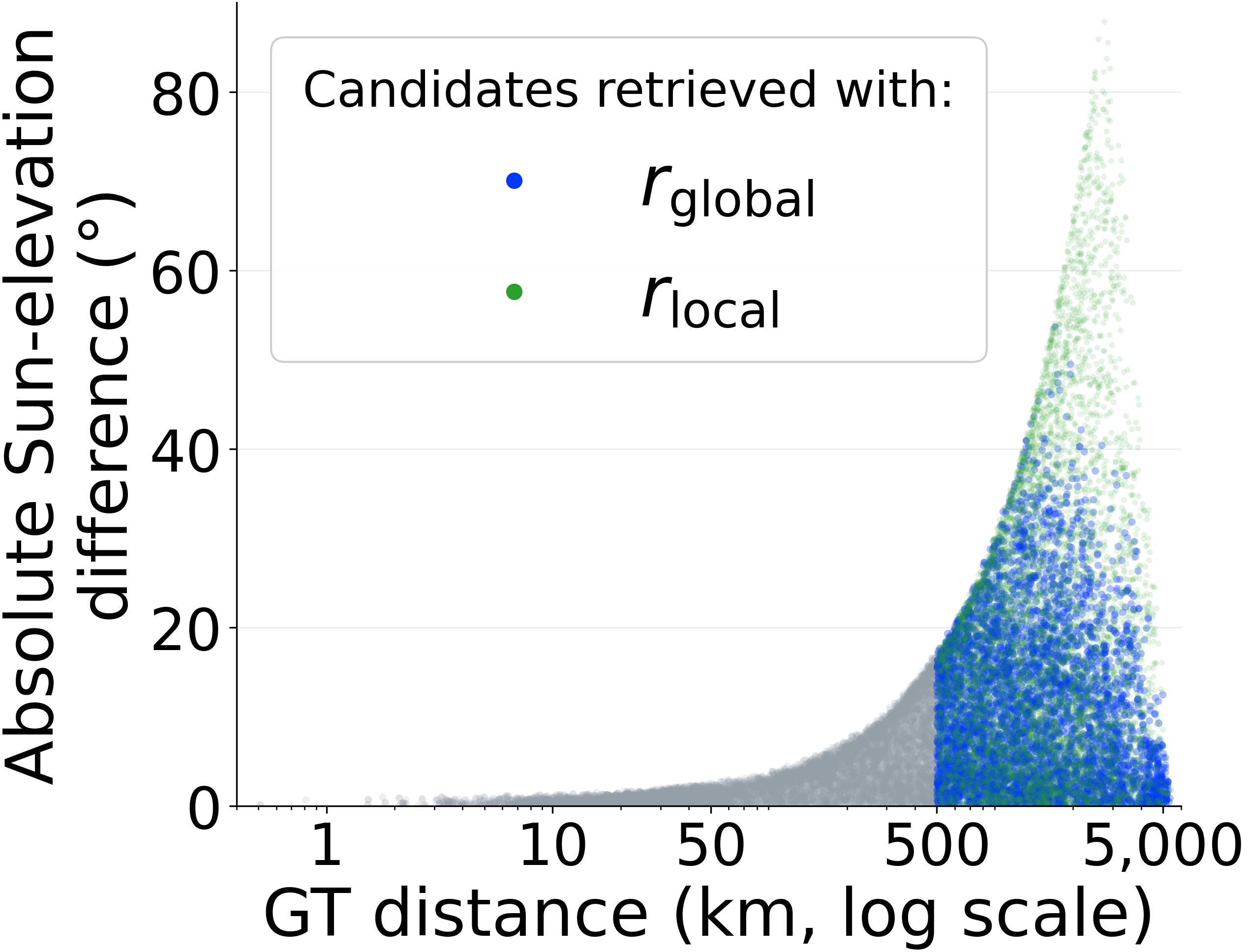}
    \hfill
    \includegraphics[width=0.49\linewidth]{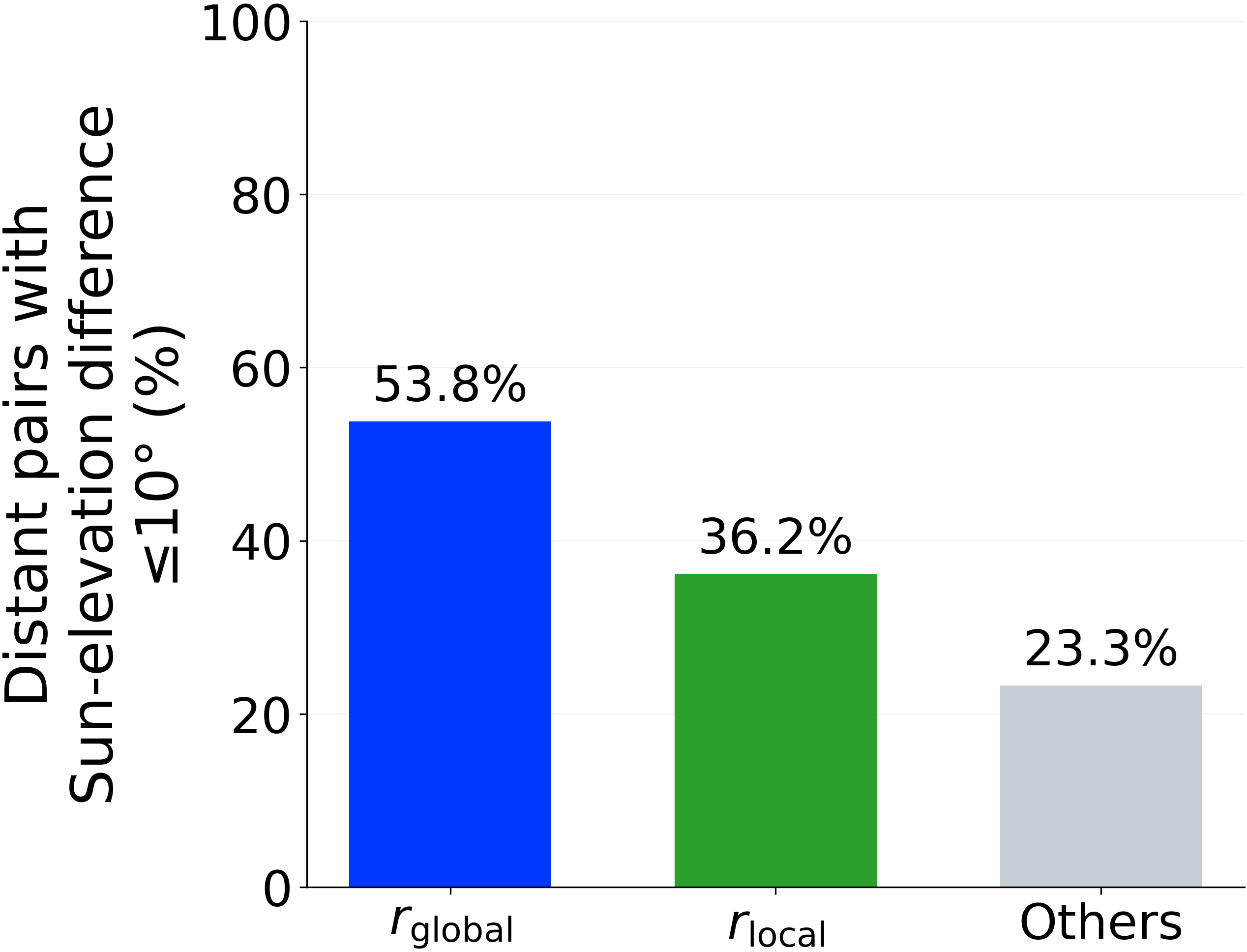}
    \vspace{-0.5em}
    \caption{\textbf{Illumination bias in DINOv2 retrieval.}
\textbf{Left:} Each point is a top-5 match. For retrievals at least 500\,km from the query, blue denotes image-level retrieval and green denotes patch-level retrieval.
\textbf{Right:} The bars show the fraction of distant matches within $10^\circ$ of the query's solar elevation, alongside the same fraction for all distant query--reference pairs.}
    \label{fig:illumination_bias}
\end{figure}

We examine candidates more than 500\,km (colored in Fig.~\ref{fig:illumination_bias}) from the query, where a retrieval is geographically incorrect. As seen in Fig.~\ref{fig:illumination_bias} (left), both scores retrieve views spanning a range of solar elevation differences, with a wider spread for patch-level matches (green) than for image-level matches (blue). Among these distant matches, $53.8\%$ of image-level matches and $36.2\%$ of patch-level matches fall within $10^\circ$ of the query's solar elevation, compared with $23.3\%$ of all reference views at those distances (Fig.~\ref{fig:illumination_bias} (right)). Similar illumination is evidently overrepresented among distant matches for the image-level score.

\vspace{-1em}
\paragraph{Effect on localization.} To test whether patch-level geometry improves localization, we vary only the ranking score within each LunarGloc-static variant, keeping its reference database and pose estimation fixed. Adding geometric agreement between matched patches to image-level similarity improves localization recall by $5.4\%$--$9.3\%$ across all three databases and all three pose-error thresholds. See supp. material for the full tabulation.

\subsection{Results including ablations}
\label{subsec:results}

\textbf{Global localization setting.} For each LunarGloc5k query, all methods use the same search region spanning
$45^\circ \times 45^\circ$ in lat-long and 30--190\,km in altitude.
When centered on the lunar equator, this region covers
$1.81 \times 10^6\,\mathrm{km}^2$ of the surface and a search volume
of $3.28 \times 10^8\,\mathrm{km}^3$.

\begin{table}[t]
\centering
\small
\setlength{\tabcolsep}{2pt}
\renewcommand{\arraystretch}{1.12}
\resizebox{\columnwidth}{!}{%
\begin{tabular}{@{}lrrrrr@{}}
\hline
\textbf{Method} & \multicolumn{3}{c}{\textbf{Recall (\%) $\uparrow$}} & \textbf{Rendered} & \textbf{Median} \\
 & \textbf{Coarse} & \textbf{Fine} & \textbf{Strict} & {\textbf{Images}$\downarrow$} & \textbf{time (s) $\downarrow$} \\
\hline
\multicolumn{6}{@{}l}{\textit{(c) Crater-based localization}} \\
STELLA~\cite{McLeod2025STELLA} & 2.29 & 1.04 & 0.18 & --- & 10.5 \\
STELLA-LFM~\cite{McLeod2025STELLA, NASAIBM2026CraterDetection} & 3.57 & 1.63 & 0.35 & --- & 13.6 \\
\hline
\multicolumn{6}{@{}l}{\textit{(a) Static database}} \\
UAVLoc-1~\cite{Wu2024UAVD4L} & 67.71 & 54.45 & 26.53 & 704 & 59.5 \\
UAVLoc-3 & 76.61 & 62.83 & 30.72 & 2,107 & 62.4 \\
UAVLoc-5 & 80.66 & 68.78 & 31.22 & 3,538 & 64.8 \\
LunarGloc-static-1 & 93.29 & 90.84 & 70.32 & 704 & 45.9 \\
LunarGloc-static-3 & 97.49 & 95.67 & 76.89 & 2,107 & 94.0 \\
LunarGloc-static-5 & \textbf{98.96} & \underline{96.96} & \underline{77.74} & 3,538 & 143.3 \\
\hline
\multicolumn{6}{@{}l}{\textit{(b) Online rendering}} \\
\multicolumn{4}{@{}l}{\textit{Mean reference images}} & 747 & \\
MCLoc~\cite{Trivigno2024MCLoc} & 65.16 & 63.27 & 55.90 & $+234$ & 1,016.5 \\
LunarGloc-w/o-guided & 87.46 & 83.10 & 60.00 & $\underline{+74}$ & 332.5 \\
LunarGloc & \underline{98.28} & \textbf{97.40} & \textbf{88.23} & $\textbf{+27}$ & 140.6 \\
\hline
\end{tabular}%
}
\caption{\textbf{Global localization results on LunarGloc5k.}
Recall uses the position and attitude error thresholds defined in
Sec.~\ref{subsec:Metrics}. For each method, we report the mean number
of images in the reference database per query. For online methods,
$+N$ denotes the mean number of additional views rendered during the search.}
\label{tab:localization-comparison}
\end{table}

\begin{figure}[t]
\centering
\includegraphics[width=\linewidth]{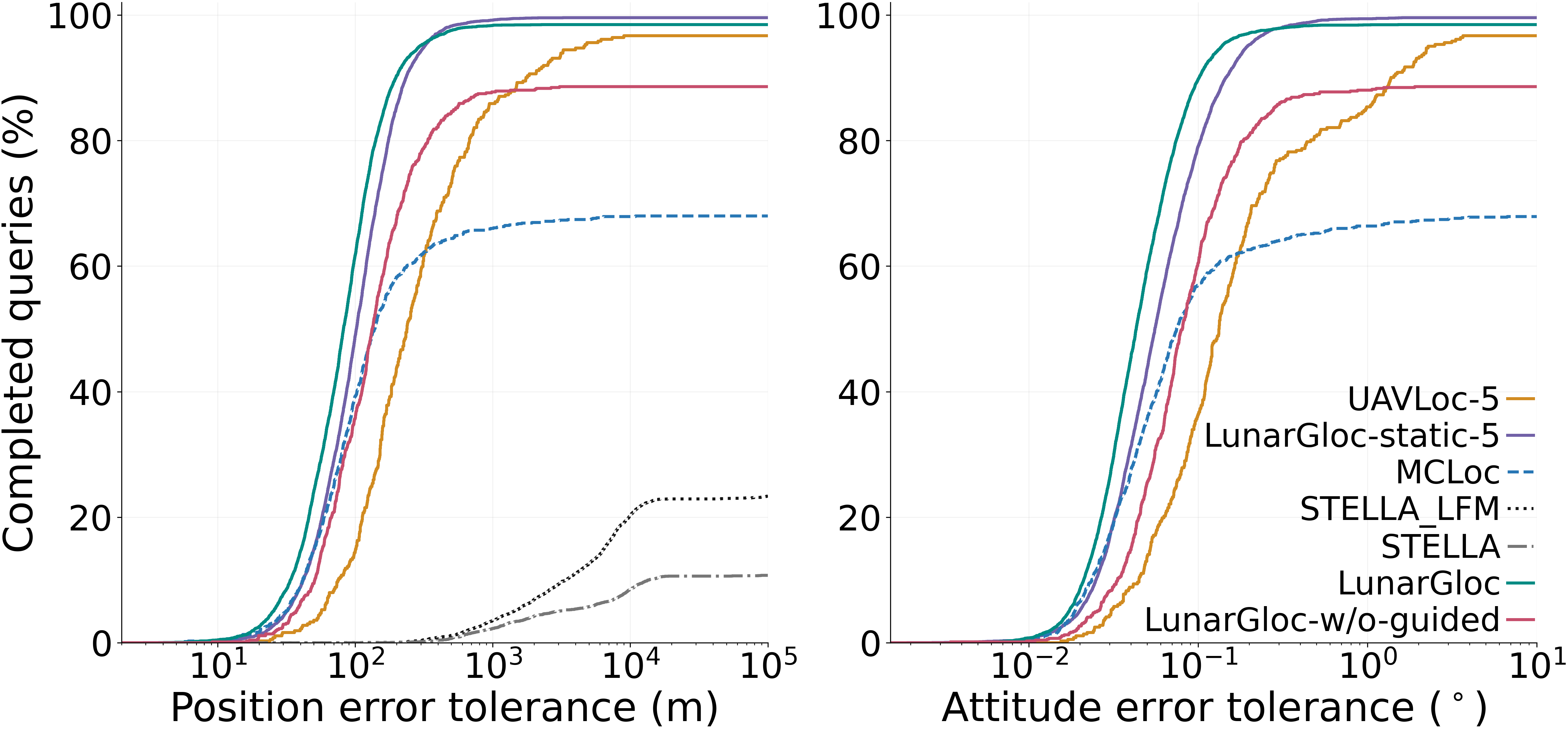}\\
\vspace{-0.5em}
\caption{Percentage of completed queries within each position (left) and attitude (right) error threshold on LunarGloc5k.}
\label{fig:cummulative_error}
\end{figure}

\vspace{-1em}
\paragraph{Comparison with crater-based methods.} The STELLA pipeline was only evaluated assuming a regional search space, $~343~\text{km}^3$~\cite{McLeod2025STELLA}. Here, we evaluated it under the much larger localization setting, which likely contributes to the much lower performance observed in Tab.~\ref{tab:localization-comparison}. 

\begin{figure}[!ht]
\centering
\includegraphics[width=0.9\linewidth]{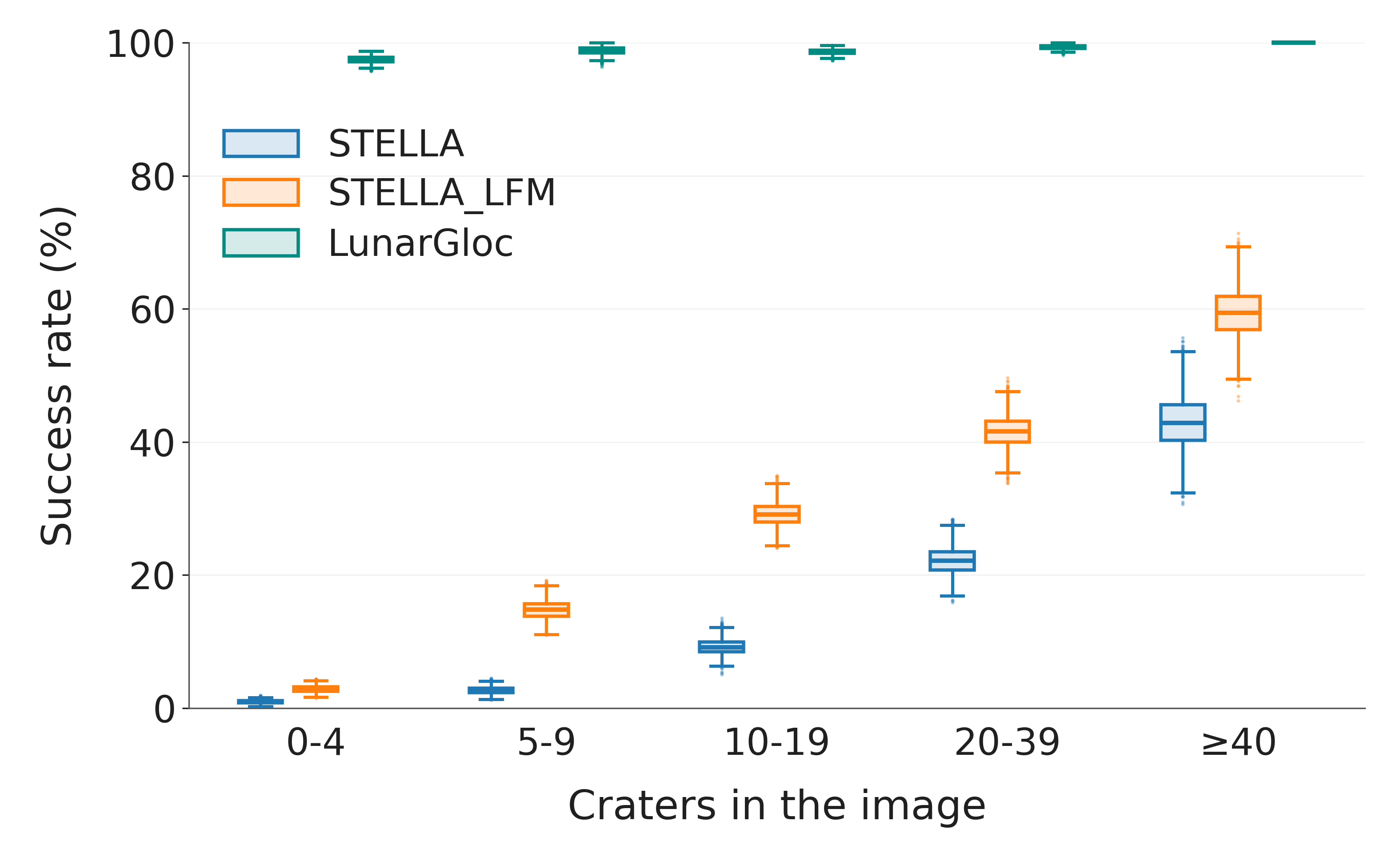}
\includegraphics[width=0.9\linewidth]{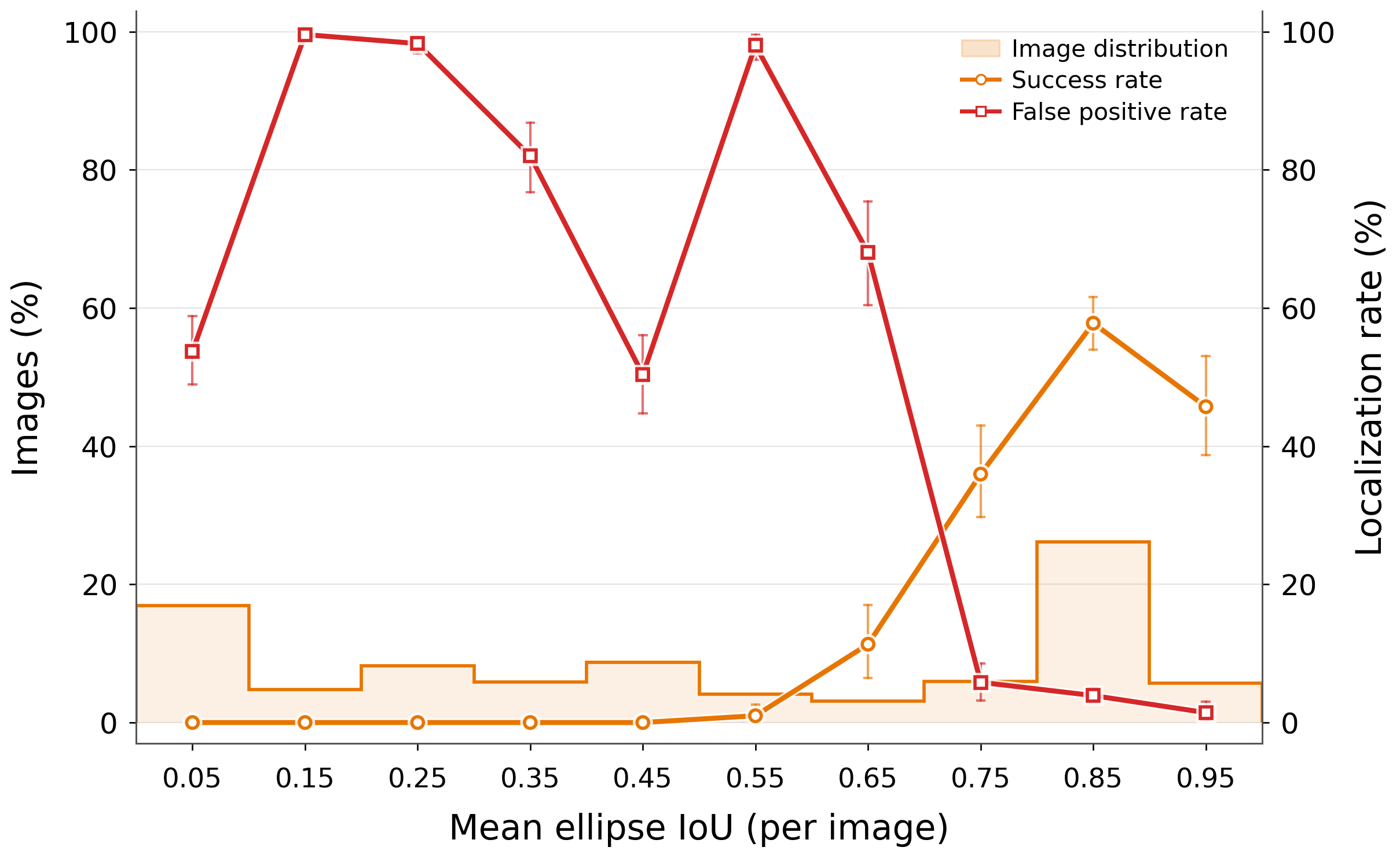}\\
\vspace{-0.5em}
\caption{\textbf{Top:} Localization success by the number of cataloged craters projected into each query image. \textbf{Bottom:} The histogram groups STELLA-LFM queries by mean ellipse IoU. The curves show the percentage in each bin with position error $\leq 10$\,km (success) or $\geq 100$\,km (false positive).}\vspace{-1.5em}
\label{fig:STELLA_vs_LunarGloc}
\end{figure}

With a more relaxed 10\,km position threshold, STELLA-LFM, the better-performing variant, achieves $20.3\%$ success rate, which is 3$\times$ better than STELLA's $8.62\%$. Fig.~\ref{fig:STELLA_vs_LunarGloc} (top) shows that the success rate increases strongly with the number of craters in the image. STELLA-LFM generally detects more craters and achieves higher overall recall. Meanwhile, the proposed method, LunarGloc, remains robust across all crater-count bins, achieving 97.5--100\% recall on completed queries under the \textit{strict} thresholds.

Another factor that strongly affects crater-based methods is the ellipse fitting quality of the detected craters. Fig.~\ref{fig:STELLA_vs_LunarGloc} (bottom) shows that only approximately $30\%$ (on average) of the images (from STELLA-LFM) have well-fitted ellipses (based on IoU) with the ground truths, which correlates strongly with the success rate. As alluded earlier, the ellipse fitting task is often harder than the detection task. Moreover, detection with poor mean IoU fit leads to high false positive rates (localization error larger than 100\,km). 

\vspace{-1em}
\paragraph{Comparison with static database methods.}
As seen Tab.~\ref{tab:localization-comparison}, adding reference altitudes improves recall for both methods. LunarGloc-static outperforms UAVLoc at every database size and threshold, but its gains diminish. Increasing coverage from one to three altitudes raises \textit{strict} recall from 70.32\% to 76.89\%. Adding two more altitudes requires another 1,431 reference images on average, and raises \textit{strict} recall by only 0.85 percentage points.

LunarGloc and LunarGloc-static-5 both exceed 98\% coarse recall, but LunarGloc achieves higher \textit{strict} recall (88.23\% vs.\ 77.74\%). We attribute this gain to online rendering, which allows LunarGloc to synthesize a view closer to the query. On average, its final candidate lies 2.50\,km from the true camera position, compared with 23.25\,km for the retrieved static view. This closer view yields more SIFT inliers for the final PnP fit (727 vs.\ 532 on average), leading to more accurate pose estimation. The supp.~material provides further details.

Apart from the discrete threshold levels, we also show the full performance profiles of the compared methods in Fig.~\ref{fig:cummulative_error}. As seen, in general, LunarGloc has the highest recall rate across the accuracy sweeps. We omit STELLA’s orientation error from Fig.~\ref{fig:cummulative_error} because it initializes the camera orientation with a small perturbation of the ground truth and restricts the orientation search to a narrow neighborhood of that initialization.

\vspace{-1em}
\paragraph{Comparison with online rendering methods.}

With a budget of up to 300 online renders per query, LunarGloc-w/o-guided achieves higher recall than MCLoc at all three thresholds, although both trail LunarGloc-static-5. Differences in search and scoring may help explain the gap between the two online methods. MCLoc randomly perturbs promising candidates, whereas LunarGloc samples from a proposal distribution that favors high-scoring regions while retaining coverage of the full search area. Secondly, DINOv2 descriptors may help LunarGloc distinguish between similar-looking lunar views more readily than the ResNet-18 descriptors used by MCLoc. 

\begin{figure}[t]
\centering
\includegraphics[width=0.85\linewidth]{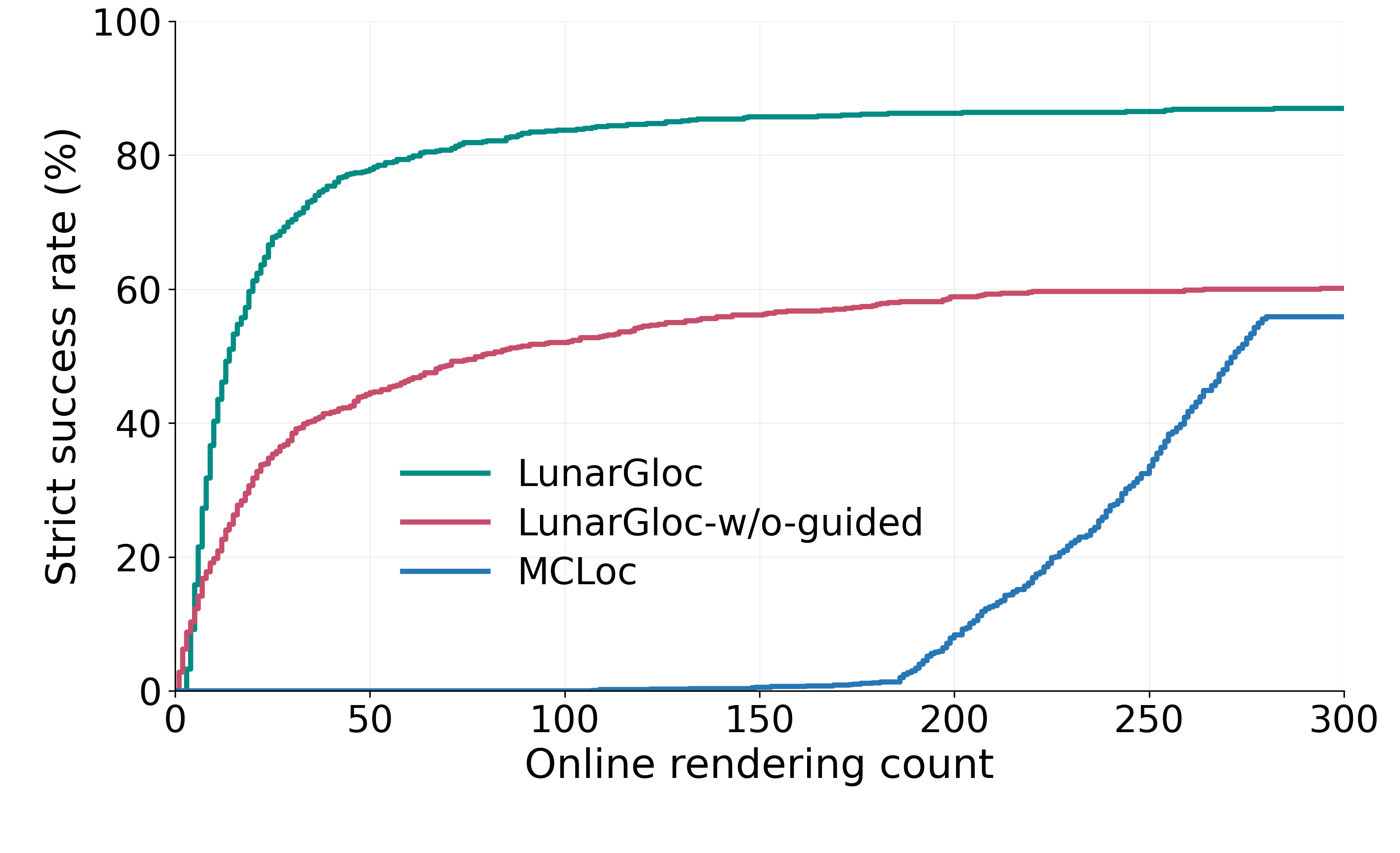}\\
\vspace{-1em}
\caption{With guided sampling, LunarGloc achieves significant gains in strict localization recall within the first 50 online renders. Its unguided variant and MCLoc require larger rendering budgets for comparable recall.}\vspace{-1em}
\label{fig:guided-render-budget}
\end{figure}

By leveraging the patch matches to guide the sampling process (Sec.~\ref{subsec:guided_sampling}), LunarGloc's \textit{strict} recall rises from 60.00\% to 88.23\%, while the mean number of online renders falls from 74 to 27 per query. The gain is already apparent at 50 renders, where LunarGloc achieves approximately 78\% \textit{strict} recall, compared with 44\% without guidance (Fig.~\ref{fig:guided-render-budget}). MCLoc's curve rises later because its hierarchical search typically selects the final view in a later refinement stage.

\vspace{-1em}
\paragraph{Runtime.}
Patch-based guidance reduces LunarGloc's median runtime from 332.5\,s
to 140.6\,s per query, compared with 1,016.5\,s for MCLoc
(Tab.~\ref{tab:localization-comparison}). LunarGloc runs in about
the same time as LunarGloc-static-5 (143.3\,s). It is substantially slower than CBN, which is a limitation of LunarGloc.

\vspace{-1em}
\paragraph{Larger-scale localization.} We evaluated LunarGloc on the larger scale localization setting. The search region is enlarged to $90^\circ \times 90^\circ$ and $180^\circ \times 180^\circ$. Note that $180^\circ \times 180^\circ$ covers the entire visible hemisphere. We observe similar performance with a moderate increase in rendering budget. Specifically, the strict success rate decreases from $88\%$ to $84\%$, while the average rendering count increases from 27 to 116, demonstrating the scalability of LunarGloc. The relevant plots are provided in the supp.~material.

\section{Conclusion}
\label{sec:conclusion}

We presented LunarGloc, a direct method for global localization of lunar orbiters from a single image. Our results show that pretrained DINOv2 descriptors can match real orbital images to rendered
views of the same lunar terrain. LunarGloc combines image-level similarity
with the geometric consistency of patch matches to score candidate views
and uses the same matches to guide its Monte Carlo search. On LunarGloc5k,
our benchmark of 5,000 real lunar orbital images, LunarGloc achieves 88.23\% recall at the strictest pose-error threshold, outperforming CBN and large-scale
VPR-based localization methods.

\vfill

\pagebreak

\section*{LLM Usage Statement}
We used OpenAI’s ChatGPT to help organize the related work, clarify the description of our method, and refine both the discussion of experimental results and the figure caption. We also used ChatGPT to help debug our code and Grammarly to refine sentence structure and phrasing.

{\small\bibliographystyle{ieeenat_fullname}\bibliography{main}}

\onecolumn
\raggedbottom
\setlength{\intextsep}{7pt plus 1pt minus 1pt}
{\centering\Large\textbf{\thetitle}\par
 \vspace{0.5em}Supplementary Material\par}
\vspace{1em}
\appendix
\setcounter{figure}{0}
\setcounter{table}{0}
\renewcommand{\theHfigure}{supp.\arabic{figure}}
\renewcommand{\theHtable}{supp.\arabic{table}}
\renewcommand{\thefigure}{S\arabic{figure}}
\renewcommand{\thetable}{S\arabic{table}}

\section*{Overview}

Sec.~\ref{sec:supp-dataset} describes the construction and annotations of
LunarGloc5k. Sec.~\ref{sec:supp-implementation} describes the renderer used
to render reference images for the static databases. Sec.~\ref{sec:supp-implementation-details} provides implementation
details, including the spacing of reference views and the hyperparameters
used for localization.

Sec.~\ref{sec:supp-scoring} extends the study of DINOv2 descriptors in
Sec.~\ref{subsec:in_depth} of the main paper by examining how changes in solar elevation
and azimuth affect descriptor similarity and retrieval.
Sec.~\ref{sec:supp-results} expands on two results from Sec.~\ref{subsec:results} of the
main paper. First, we show how the distance between each
LunarGloc-static variant's final candidate and the ground-truth query
position relates to pose accuracy after refinement. Second, we provide
the full plot of LunarGloc's strict recall and online renders per query
as the search region expands to a visible lunar hemisphere. Finally,
Sec.~\ref{sec:supp-qualitative} presents qualitative comparisons of
LunarGloc and its variants with competing methods, highlighting common
failure cases of the baselines.

\section{LunarGloc5k construction and annotations}
\label{sec:supp-dataset}

\subsection{Source imagery and annotations}
\label{sec:supp-sources}
LunarGloc5k uses calibrated 643\,nm imagery from the Wide Angle Camera (WAC) aboard NASA's Lunar Reconnaissance Orbiter (LRO)~\cite{Robinson2009LROCCDR,Robinson2010LROC}. In visible-color mode, WAC spans approximately $60^\circ$ across track. WAC is chosen due to its wide field of view, which resembles the typical camera setting for an orbiter's localization module. 

\begin{figure}[H]
  \centering
  \includegraphics[width=\linewidth]{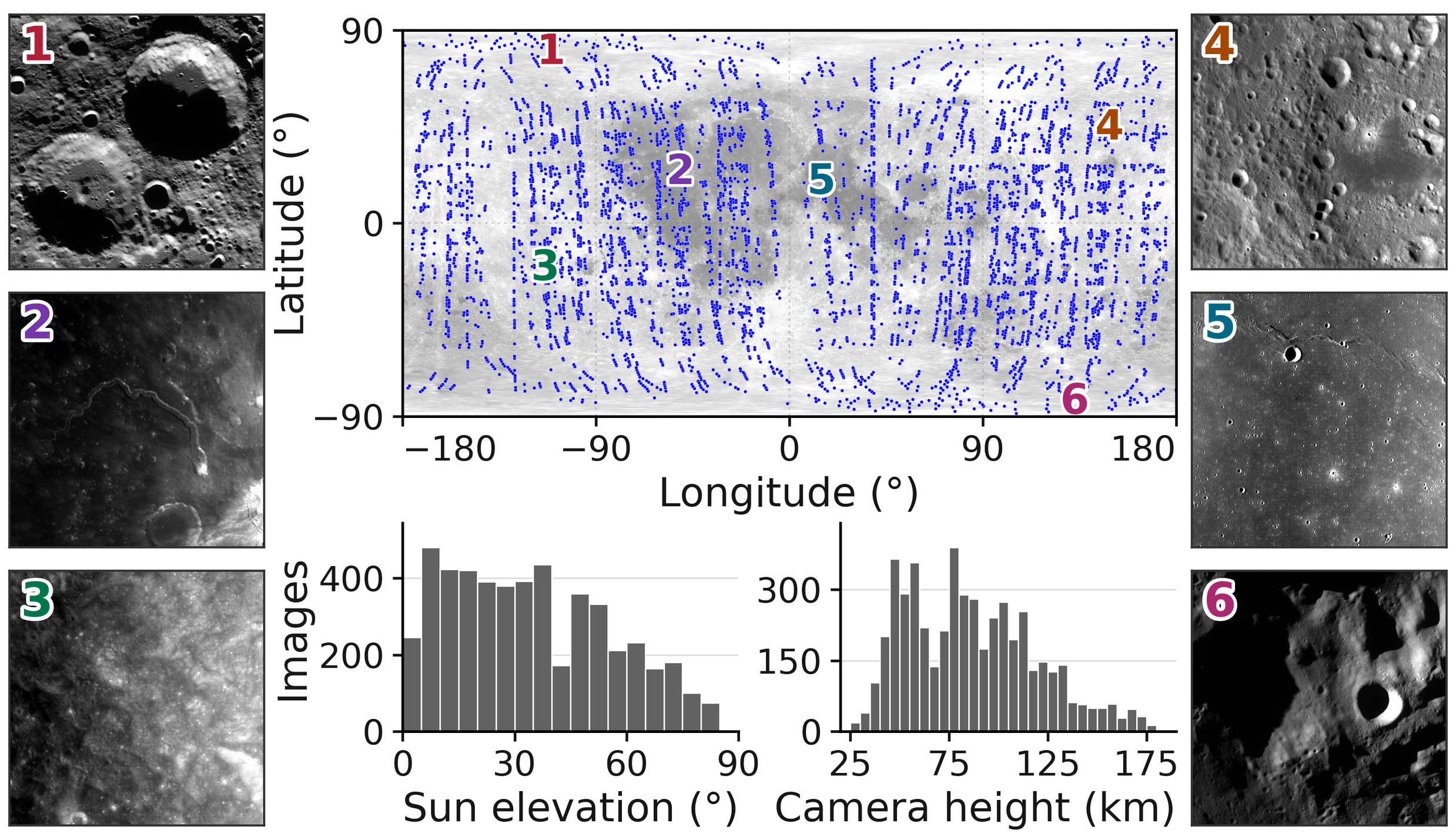}
  \caption{\textbf{Geographic coverage and viewing conditions of LunarGloc5k.}
Blue points mark all 5,000 query locations. The six numbered images correspond
to the marked sites. Sites 2 and 5 show lunar maria with few visible craters,
while the north and south polar views (1 and 6) with low local sun elevation angles.
The histograms show the distributions of solar elevation (left) and camera
height above the lunar reference sphere (right). Lunar basemap courtesy of
NASA's Scientific Visualization Studio.}
  \label{fig:lunarloc5k-coverage}
\end{figure}

We selected 5,000 distinct anchor framelets from 1,312 WAC products acquired over 377 LRO orbits between 11 October 2009 and 24 April 2026. Each anchor defines one $480\times480$-pixel perspective query with a $60^\circ\times60^\circ$ field of view (FOV). For each query, we record the virtual camera pose and intrinsics, the source product and framelet identifiers, and the acquisition time in UTC. Sec.~\ref{sec:data} of the main paper illustrates how the source framelets form a query. Fig.~\ref{fig:lunarloc5k-coverage} shows the geographic coverage, the distributions of solar elevation and camera altitude, and example query images.

\subsection{Framelet reprojection}
\label{sec:supp-framelets}
For each anchor framelet, we use neighboring framelets from the same
WAC product. We backproject their calibrated intensities onto the lunar
DEM using the SPICE camera geometry at each framelet's exposure-center
time. We then sample the resulting intensity map from a virtual pinhole
camera centered at the anchor's physical camera center, with its optical
axis aligned to the anchor's calibrated center ray. The pose label
describes this virtual camera, and its accuracy depends on the SPICE
geometry and the DEM. This procedure is geometrically analogous to
the camera-to-camera reprojection in ISIS's \texttt{cam2cam}
tool~\cite{ISISCam2Cam}.

\clearpage

\section{Renderer}
\label{sec:supp-implementation}
\subsection{Lunar terrain and surface reflectance}
\label{sec:supp-renderer}
\paragraph{Terrain mesh.} We construct the lunar terrain mesh in Blender using the SLDEM2015~\cite{Barker2016SLDEM} at 512 pixels per degree between $60^\circ$S and $60^\circ$N, and the 60\,m/pixel LOLA DEMs~\cite{Neumann2010LOLAGDR,Smith2010LOLA} at higher latitudes. We convert each elevation sample to a Moon-centred Cartesian vertex and split each grid cell into two triangles.

\paragraph{Surface reflectance.}
We use the 643\,nm WAC\_EMP mosaic~\cite{LROCWACEmpiricalMosaic} as the
surface texture at all latitudes. For views centered between $60^\circ$S
and $60^\circ$N ($|\phi| < 60^\circ$), we model surface reflectance using
the local Hapke parameters of Sato et al.~\cite{Sato2014HapkeMaps}.
At $|\phi| \geq 60^\circ$, we use a diffuse reflectance model with the
same texture.

\subsection{Visual gap reduction with the temporal information}
\label{subsec:visual_gap}
\begin{figure}[H]
  \centering
  \includegraphics[width=0.89\linewidth]{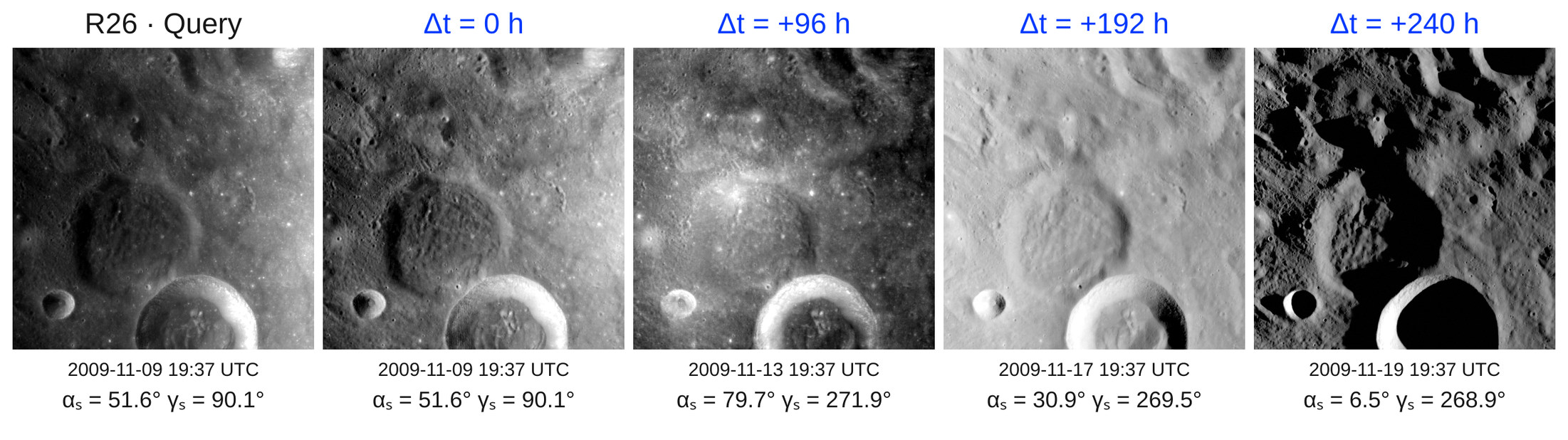}\par\smallskip
  \caption{\textbf{Effect of illumination on the appearance of the same lunar terrain.}
The real query image (left) and four renders from the same camera pose.
Rendering at the query timestamp ($\Delta t = 0$) reproduces the observed
crater shading. As time elapses after query acquisition, solar elevation ($\alpha_{s}$) and azimuth ($\gamma_{s}$) change.
At $+96$\,h, crater rims appear less distinct. By $+240$\,h, they cast long shadows.}
  \label{fig:supp-illumination-examples}
\end{figure}

As described in Eq.~\eqref{eq:opt} in Sec.~\ref{subsec:global-localization}, the renderer takes in the query timestamp, which is typically obtained using the onboard clock. For example, LRO WAC image products record both the spacecraft clock count at the start of acquisition and the corresponding UTC time~\cite{Robinson2009LROCCDR}. This allows us to reduce the visual gap caused by illumination. Fig.~\ref{fig:supp-illumination-examples} visualizes the drastic visual change of the same location at different acquisition times. We further demonstrate in Sec.~\ref{sec:supp-scoring} the effect of these illumination effects on DINOv2' descriptors. 

\section{Implementation details}
\label{sec:supp-implementation-details}

\subsection{Reference database construction}
\label{sec:supp-database}
For the reference database, we adopt the same camera orientation assumption as stated in Sec.~\ref{subsec:global-localization}. It simulates a downward-looking camera that covers the search surface. The reference view has the same intrinsic parameters (480 $\times$ 480 pixels, $60^\circ \times 60^\circ$ FOV) as the LunarGloc5k dataset (which we stated in the main paper as the assumed known parameter in the search). It covers approximately 58\,km along each axis of the lunar surface.

For each altitude, we target a 20\% overlap between adjacent views, giving a nominal
center-to-center spacing of 46.4\,km. The altitude layers described in Sec.~\ref{subsec:baseline} of the main paper reuse the same horizontal grid, so each additional layer captures the same locations at a different image scale.

\subsection{Localization hyperparameters}
\label{sec:supp-hyperparameters}
Tab.~\ref{tab:supp-hyperparameters} supplements Sec.~\ref{subsec:hyper} in the main paper, which lists LunarGloc's hyperparameters used for the standard $45^\circ \times 45^\circ$ search. The SIFT matching, pose refinement and pose acceptance hyperparameters also apply to LunarGloc-static, LunarGloc-w/o-guided, and MCLoc.

\begin{table}[H]
\centering
\small
\caption{LunarGloc hyperparameters for the standard
$45^\circ \times 45^\circ$ search. Pixel thresholds are measured
at the original image resolution.}
\label{tab:supp-hyperparameters}
\begin{tabular}{@{}p{0.18\linewidth}p{0.45\linewidth}p{0.28\linewidth}@{}}
\toprule
Stage & Parameter & Value \\
\midrule
Global search
  & Adaptive mixture weight (Eq.~\eqref{eq:q_n} in the main paper) $\beta$ & 0.75 \\
  & Spherical kernel bandwidth (Eq.~\eqref{eq:guided-density-update} in the main paper) & 20\,km \\
  & Altitude range for random samples & 40--160\,km \\
  & Maximum online renders per query & 300 \\
\midrule
Patch matching
  & Inlier tolerance & 1 patch spacing ($1/16$ of image width) \\
  & Minimum inliers ($\vert \mathcal{C}^* \vert$) to invoke guided search (Sec.~\ref{subsec:guided_sampling}) & 8 \\
  & Minimum inliers ($\vert \mathcal{C}^* \vert$) to invoke pose refinement (Sec.~\ref{subsec:pnp}) &  25 \\
\midrule
SIFT matching
  & Maximum keypoints per image & 10{,}000 \\
  & Contrast threshold / nearest-neighbor ratio threshold & 0.015 / 0.72 \\
\midrule
Pose refinement
  & RANSAC reprojection threshold & 4\,px \\
  & Maximum iterations / confidence & 10{,}000 / 0.9999 \\
\midrule
Pose acceptance
  & Minimum PnP inliers / minimum inlier ratio & 8 / 5\% \\
  & Maximum RMS inlier reprojection error & 4\,px \\
  & Minimum fraction of points with positive depth & 95\% \\
  & Minimum query image area covered by the inlier convex hull & 0.2\% \\
\bottomrule
\end{tabular}
\end{table}

For the larger search regions in Fig.~\ref{fig:larger-scale_loc} and in Sec.~\ref{subsec:results} (\textbf{Larger-scale localization}), all queries in the $90^\circ \times 90^\circ$ experiment used
the same 300-render budget. In the $180^\circ \times 180^\circ$
experiment, we increase the budget to 1,000 images.

\clearpage

\section{DINOv2 scoring under illumination changes}
\label{sec:supp-scoring}
\begin{figure}[H]
  \centering
  \includegraphics[width=0.89\linewidth]{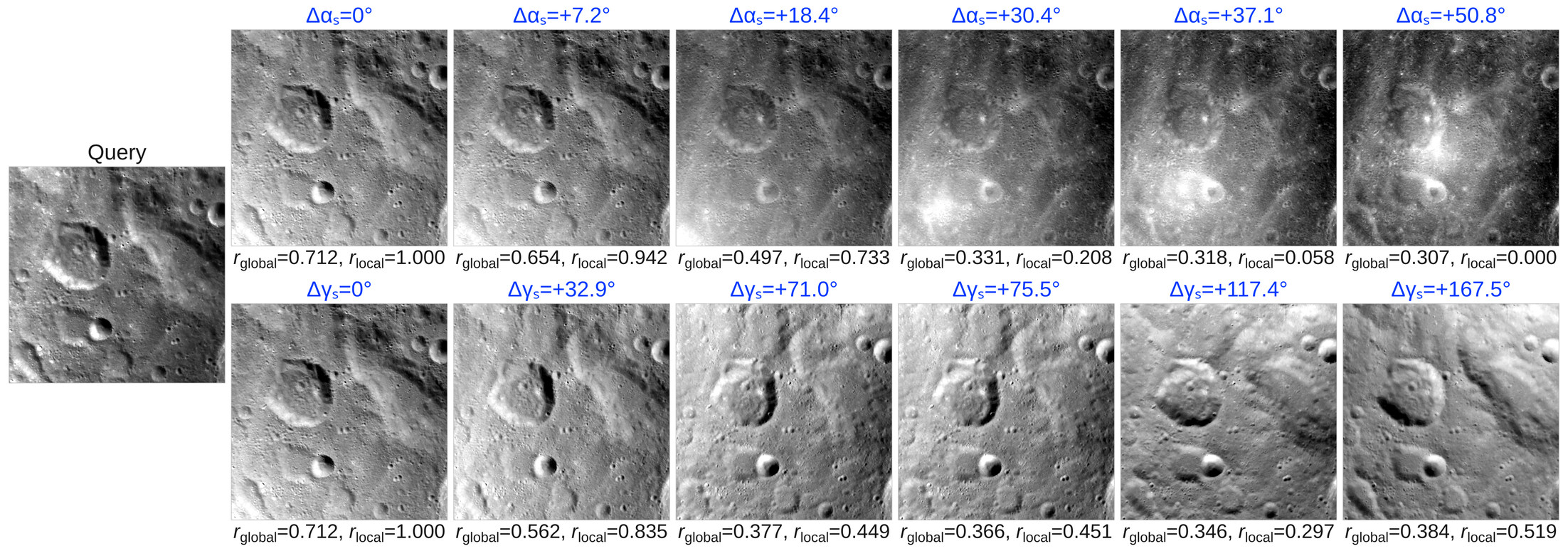}\par\smallskip
  \includegraphics[width=0.89\linewidth]{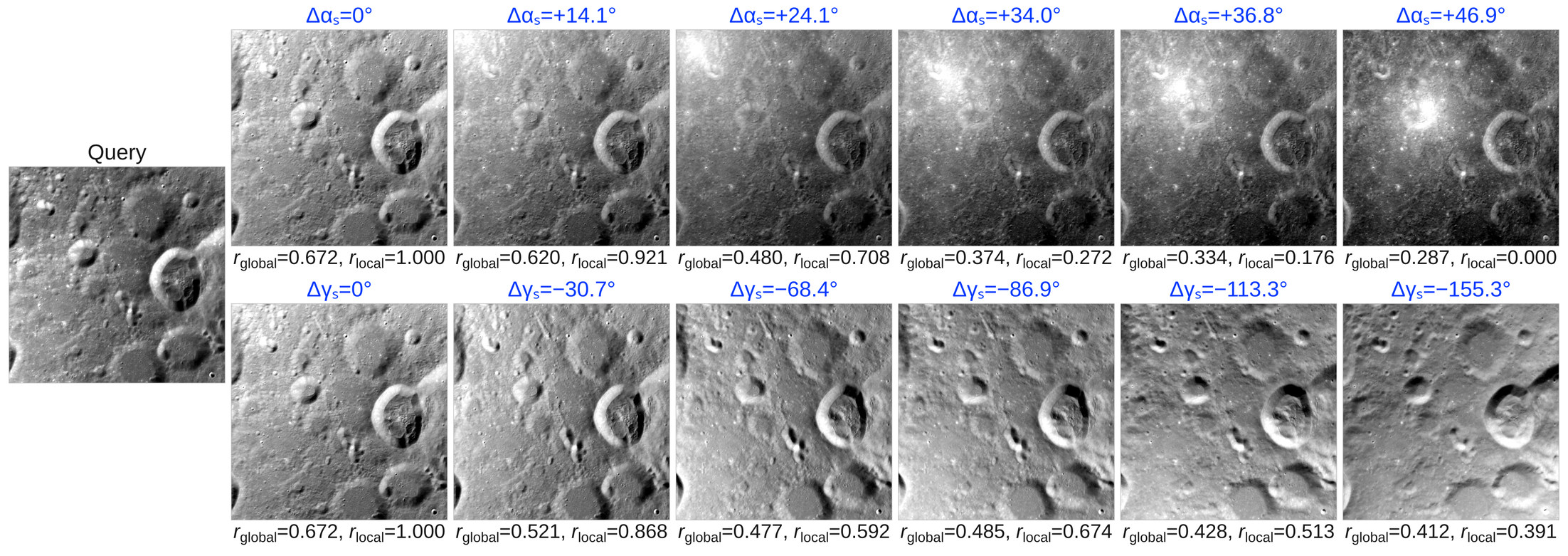}
\caption{\textbf{Effect of solar angle deviations on query--render similarity.}
For each real query (left), we render the same terrain at a fixed camera pose
while varying either solar elevation (upper row) or azimuth (lower row).
Larger deviations from the query's illumination generally yield lower
similarity, as indicated by the global and local matching scores
$r_{\mathrm{global}}$ and $r_{\mathrm{local}}$.}  
  \label{fig:supp-illumination-examples-for-study}
\end{figure}
\subsection{Sensitivity to solar elevation and azimuth}
\label{sec:supp-illumination}
As alluded to in Sec.~\ref{subsec:visual_gap}, we investigated the sensitivity of DINOv2's descriptors to the illumination condition. 

\begin{figure}[H]
  \centering
  \includegraphics[width=0.445\linewidth]{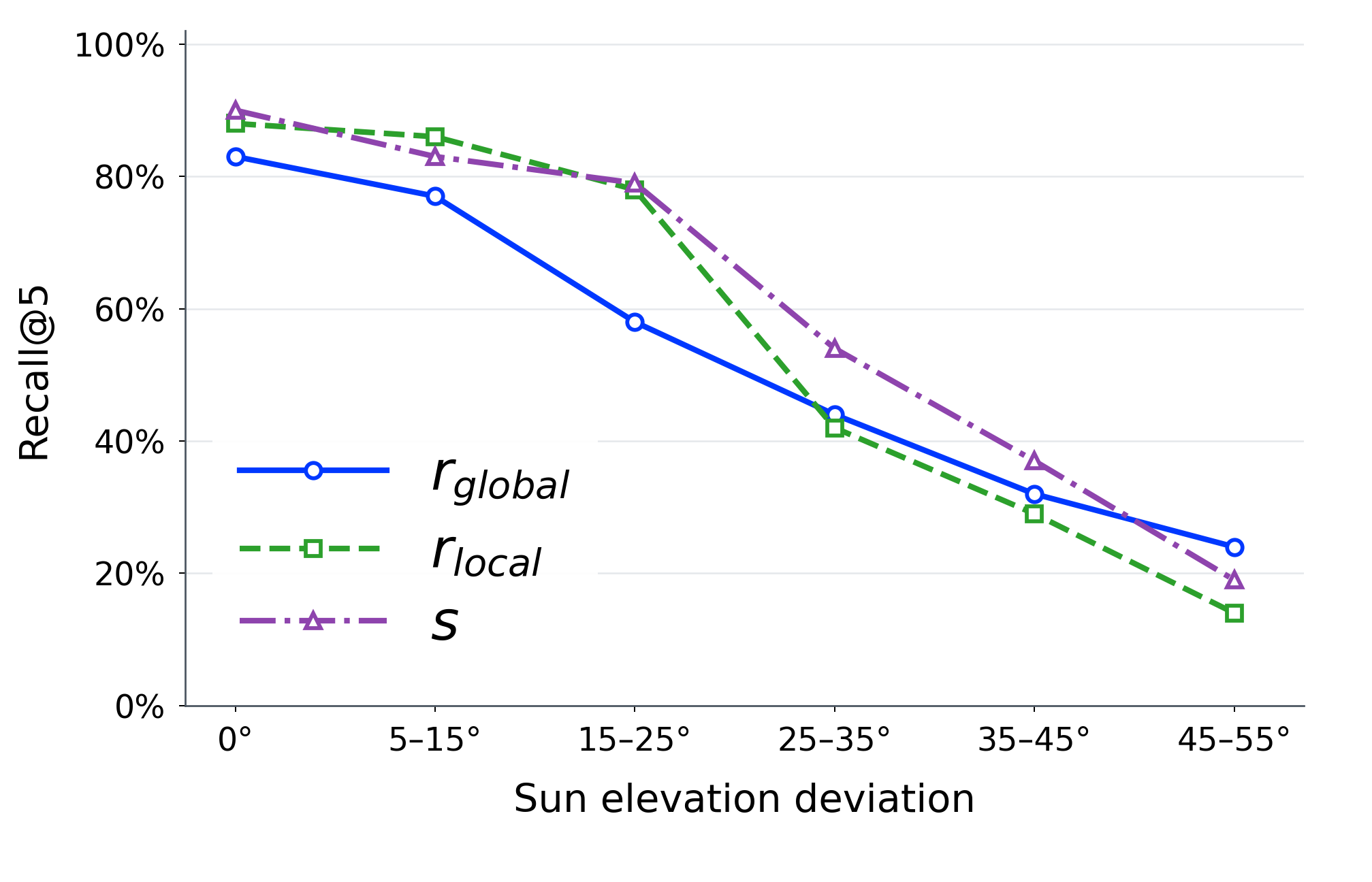}\hspace{0.035\linewidth}%
  \includegraphics[width=0.445\linewidth]{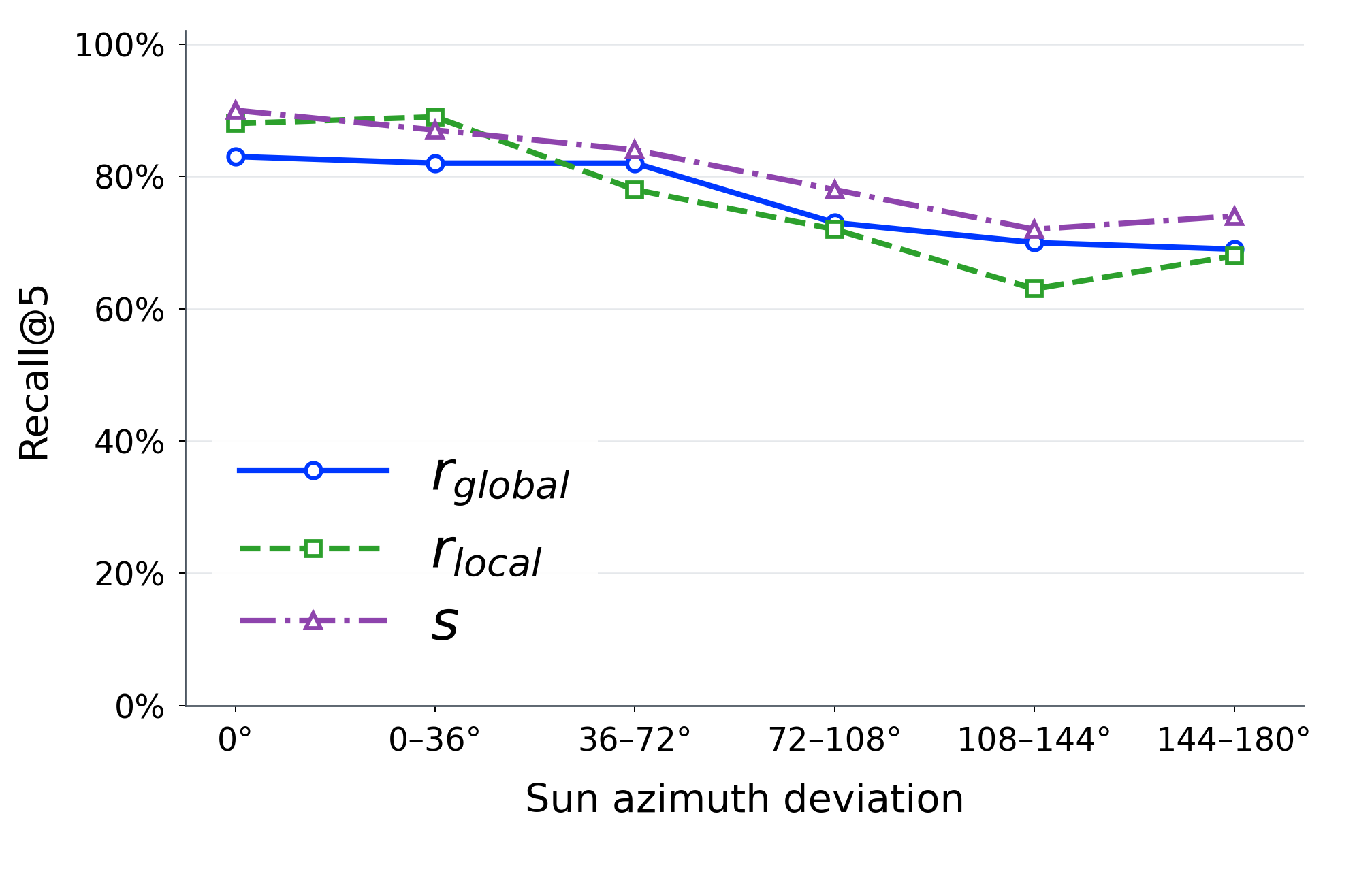}
\caption{\textbf{Retrieval recall by solar angle offset.} We grouped rendered references into bins by the magnitude of their offset from the query's solar elevation (left) or azimuth (right), as described in \textbf{Setup}. Each bin forms a reference database, and the $0^\circ$ bin matches the
query illumination. Each point reports Recall@5 over 100 real query images
using the indicated score. Recall declines more sharply for elevation
offsets, with the global score $r_{\mathrm{global}}$ dropping earlier
than the local score $r_{\mathrm{local}}$ and the combined score $s$.}
  \label{fig:supp-illumination-recall}
\end{figure}

\paragraph{Setup.} We vary solar elevation and azimuth
independently in the following steps.
\begin{enumerate}
    \item \textbf{Select the reference poses.}
    For each of 100 real queries in LunarGloc5k, we select the spatially closest reference
    view in the database. We keep its camera pose fixed across all renders
    for that query.

    \item \textbf{Vary one solar angle at a time.}
    In the elevation experiment, we render each site at five offsets (at varying magnitude) from
    the query's solar elevation while keeping azimuth fixed at the query
    value. In the azimuth experiment, we reverse the roles of the two
    angles. Each experiment includes a zero-offset control that matches
    the query illumination. This gives six databases per angle, each
    containing one render per site. Fig.~\ref{fig:supp-illumination-examples-for-study} shows two examples.

    \item \textbf{Measure retrieval recall.}
    For each query, we rank the 100 references in each database separately
    using the global score $r_{\mathrm{global}}$, the local score
    $r_{\mathrm{local}}$, and the combined score $s$. Fig.~\ref{fig:supp-illumination-recall} reports Recall@5, the percentage of queries whose corresponding reference
    ranks among the top five.
\end{enumerate}

\paragraph{Observations.}
Recall@5 generally decreases as either solar elevation or azimuth deviates
from its value at query acquisition (Fig.~\ref{fig:supp-illumination-recall}). The decline is more significant for elevation, consistent with the larger visual changes as shown in Fig.~\ref{fig:supp-illumination-examples-for-study}.

The left panel of Fig.~\ref{fig:supp-illumination-recall} also shows that the choice of retrieval score matters. As the elevation offset increases, recall drops more sharply for
the global score $r_{\mathrm{global}}$ than for the local score
$r_{\mathrm{local}}$ or the combined score $s$. This greater sensitivity
of global retrieval to illumination is consistent with both studies in
Sec.~\ref{subsec:in_depth} of the main paper.

\subsection{Effect of patch geometry on localization recall}
\label{sec:supp-score-ablation}
Tab.~\ref{tab:global-only-vs-fused} supplements the results of the second study in Sec.~\ref{subsec:in_depth} of the main paper. It is an ablation study on the impact of using different scores on the localization result. 

\paragraph{Setup.} For each LunarGloc-static
database, we rank the same references using the global score
$r_{\mathrm{global}}$, the local score $r_{\mathrm{local}}$, or their
combination $s$. The three highest-ranked references then undergo the same
pose refinement. Alongside the standard 100\,km single-layer database and
the three- and five-layer databases, we evaluate the 50\,km single-layer
database used to initialize LunarGloc.

\paragraph{Observations.}
All three retrieval scores yield lower localization recall with the 50\,km
database than with the 100\,km database (Tab.~\ref{tab:global-only-vs-fused}).
We attribute this gap in part to image scale. Query altitudes span
25--190\,km (Fig.~\ref{fig:lunarloc5k-coverage}), and 61\% of queries are
closer in altitude to the 100\,km layer. For these queries, the 50\,km
references cover a smaller footprint and show terrain features at a larger
scale. LunarGloc uses the 50\,km database because it also
serves as the source imagery for rendering by projection (Sec.~\ref{subsec:renderer} in the main paper). The finer resolution at 50\,km avoids the need to upsample
the reference images (which leads to artifacts) when rendering the majority of the higher-atltitude candidate views within the search range (40km - 160km). 

The combined score $s$ achieves the highest recall at all three thresholds
for both single-layer databases. Its advantage is clearest for the 50\,km
database, where coarse recall reaches 77.44\%, compared with 69.33\% for
$r_{\mathrm{global}}$ and 69.48\% for $r_{\mathrm{local}}$. This gain
motivates using $s$ to initialize LunarGloc's proposal density from the
50\,km database. As more altitude layers are added, the gap between
$r_{\mathrm{local}}$ and $s$ narrows. With five layers, their coarse and
fine recall are nearly identical, and $r_{\mathrm{local}}$ achieves slightly
higher strict recall (78.33\% vs.\ 77.74\%).

\begin{figure}[H]
  \centering
  \includegraphics[width=0.93\linewidth]{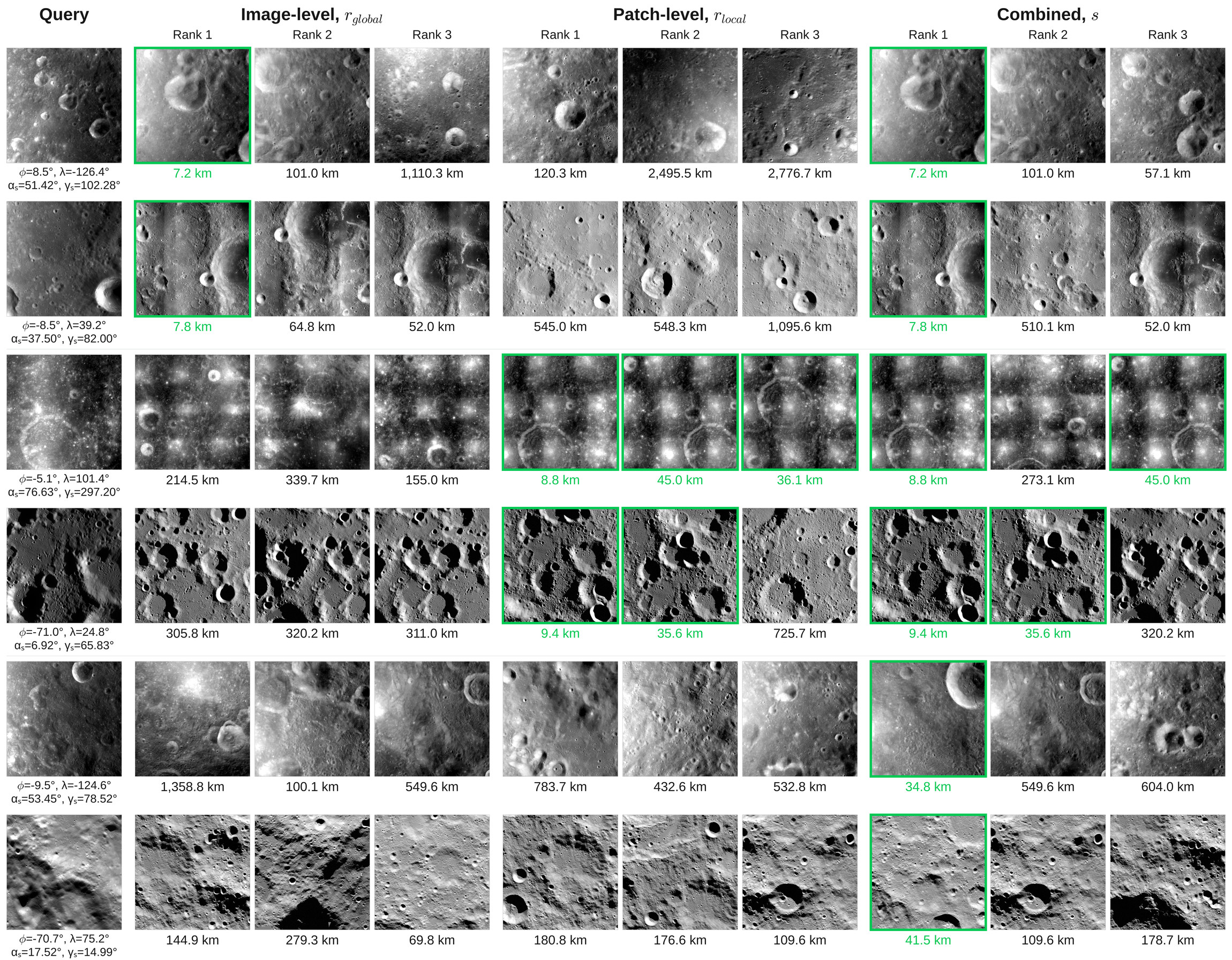}
  \caption{\textbf{Top-3 reference retrievals using $r_{global}$, $r_{local}$, and $s$.} Each row shows a query followed by the three highest-ranked references
for each score. Labels beneath the references give the distance from each
reference center to the ground-truth query location. Green outlines mark
references within 50\,km. The combined score ranks such a reference first
in all six examples. In the last two rows, both individual scores miss it
entirely in their top three.}
  \label{fig:global_local_combined}
\end{figure}

We show qualitative retrieval examples from each method in Fig.~\ref{fig:global_local_combined}. As seen in the first three columns after query, the $r_{global}$ tends to retrieve candidates based on the illumination condition, which is the dominant visual cue in the image. As seen in the forth and sixth rows, it also tends to retrieve similar-structure scenes that are far away from the query.

Meanwhile, the most obvious evidence that $r_{local}$ is less sensitive to the illumination cue is in the second and fifth rows, where the top-3 retrievals have different illumination than the query. Its strength is portrayed in the third and forth rows, where it pays attention to the local features and retrieves the correct candidates. In all these cases, we show that the combination score $s$ is the better candidate, as it manages to retrieve the correct candidates. 

\begin{table}[H]
\centering
\footnotesize
\setlength{\tabcolsep}{3pt}
\renewcommand{\arraystretch}{1.08}
\begin{tabular*}{0.76\linewidth}{@{\extracolsep{\fill}}llrrr@{}}
\hline
\textbf{Method} & \textbf{Ranking} & \multicolumn{3}{c}{\textbf{Recall (\%) $\uparrow$}} \\
 & & \shortstack{\textbf{Coarse}\\$\leq1$\,km, $\leq0.5^\circ$} & \shortstack{\textbf{Fine: }\\$\leq600$\,m$, \leq0.25^\circ$} & \shortstack{\textbf{Strict: }\\$\leq300$\,m, $<0.1^\circ$} \\
\hline
LunarGloc-static-1 (50km) & $r_{\mathrm{global}}$ & 69.33 & 59.02 & 28.61 \\
 & $r_{\mathrm{local}}$ & 69.48 & 62.47 & 31.80 \\
 & $s$ & \textbf{77.44} & \textbf{67.48} & \textbf{32.80} \\
\hline 
LunarGloc-static-1 (100km) & $r_{global}$ & 84.68 & 81.52 & 63.75 \\
 & $r_{local}$ & 91.63 & 89.39 & 69.20 \\
 & $s$ & \textbf{93.29} & \textbf{90.84} & \textbf{70.32} \\
\hline
LunarGloc-static-3 & $r_{global}$ & 89.17 & 87.37 & 71.45 \\
& $r_{local}$ & 97.29 & 95.49 & 76.85 \\
  & $s$ & \textbf{97.49} & \textbf{95.67} & \textbf{76.89} \\
\hline
LunarGloc-static-5 & $r_{global}$ & 91.34 & 88.65 & 70.41 \\
& $r_{local}$ & \textbf{98.94} & 96.94 & \textbf{78.33} \\
 & $s$ & \textbf{98.94} & \textbf{96.96} & 77.74 \\
\hline
\end{tabular*}
\caption{\textbf{Effect of the retrieval score on LunarGloc-static localization recall.}
For each reference database, we vary the score used to select the top three
references for pose refinement. The best recall rate for each variant are bolded.}
\label{tab:global-only-vs-fused}
\end{table}

\clearpage

\section{Additional localization results}
\label{sec:supp-results}

\subsection{Converged candidate and pose refinement}
\label{sec:supp-candidates}
We highlighted that LunarGloc outperforms the LunarGloc-static baseline at the strictest level threshold because it converges to a much closer candidate before the final pose refinement (Sec.~\ref{subsec:pnp} in the main paper), which performs SIFT feature matching and PnP. 

Tab.~\ref{tab:candidate-quality} compares the final candidates on queries for which all four methods returned a pose. For the methods that use a static database, the distance between the retrieved candidates and the query image decreases, from 38 km to 23.25 km on average, as the database resolution increases. Naturally, this leads to a higher SIFT matches count (increased from 416 to 532) and eventually higher recall rates (increased from 70.32$\%$ to 77.74$\%$). 

\begin{figure}[H]
\centering
\includegraphics[width=\textwidth]{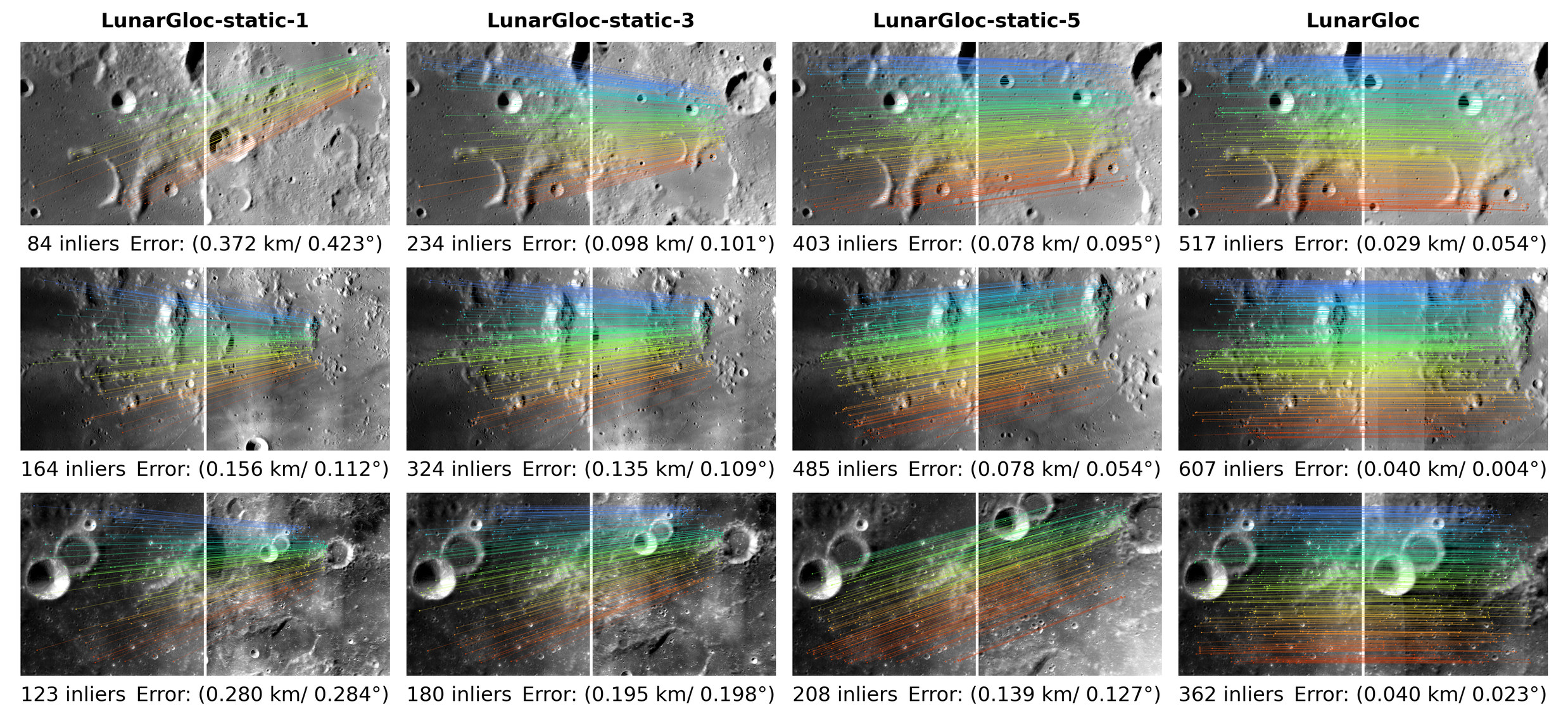}
\caption{\textbf{SIFT matches used for pose refinement.}
Each row shows one query paired with the final reference selected by each
method. Lines connect inlier matches, and annotations give the inlier count
and final position and attitude errors. The closer the final candidate is to the query image, the higher the SIFT inlier count is, which leads to higher pose accuracy.}
\label{fig:sift-matches-example}
\end{figure}

LunarGloc performs online rendering, which typically converges to the final candidate much closer to the query, around 2.5 km on average (more than $9\times$ better than the LunarGloc-static-5), which leads to the highest SIFT inliers (727) and recall rate (88.23) among the competition. Fig.~\ref{fig:sift-matches-example} illustrates some examples of the final retrieved candidates and their SIFT matches. 

\begin{table}[H]
\centering
\footnotesize
\setlength{\tabcolsep}{2pt}
\renewcommand{\arraystretch}{1.05}
\begin{tabular*}{0.76\linewidth}{@{\extracolsep{\fill}}lrrr@{}}
\hline
\textbf{Method} & \shortstack[r]{\textbf{Mean candidate}\\\textbf{distance (km)}} & \shortstack[r]{\textbf{Mean SIFT}\\\textbf{inliers}} & \shortstack[r]{\textbf{Strict}\\\textbf{acc. (\%)}} \\
\hline
LunarGloc-static-1 & 38.04 & 416 & 70.32 \\
LunarGloc-static-3 & 26.43 & 510 & 76.89 \\
LunarGloc-static-5 & 23.25 & 532 & 77.74 \\
LunarGloc & \textbf{2.50} & \textbf{727} & \textbf{88.23} \\
\hline
\end{tabular*}
\caption{\textbf{Final reference quality and pose refinement.}
We report the mean distance from the selected reference to the ground-truth
query position, the mean number of SIFT inliers, and the percentage of queries
meeting the strict pose threshold after refinement.}
\label{tab:candidate-quality}
\end{table}

\subsection{Larger-scale localization}
\label{sec:supp-scale}
\begin{figure}[H]
\centering
\includegraphics[width=0.58\linewidth]{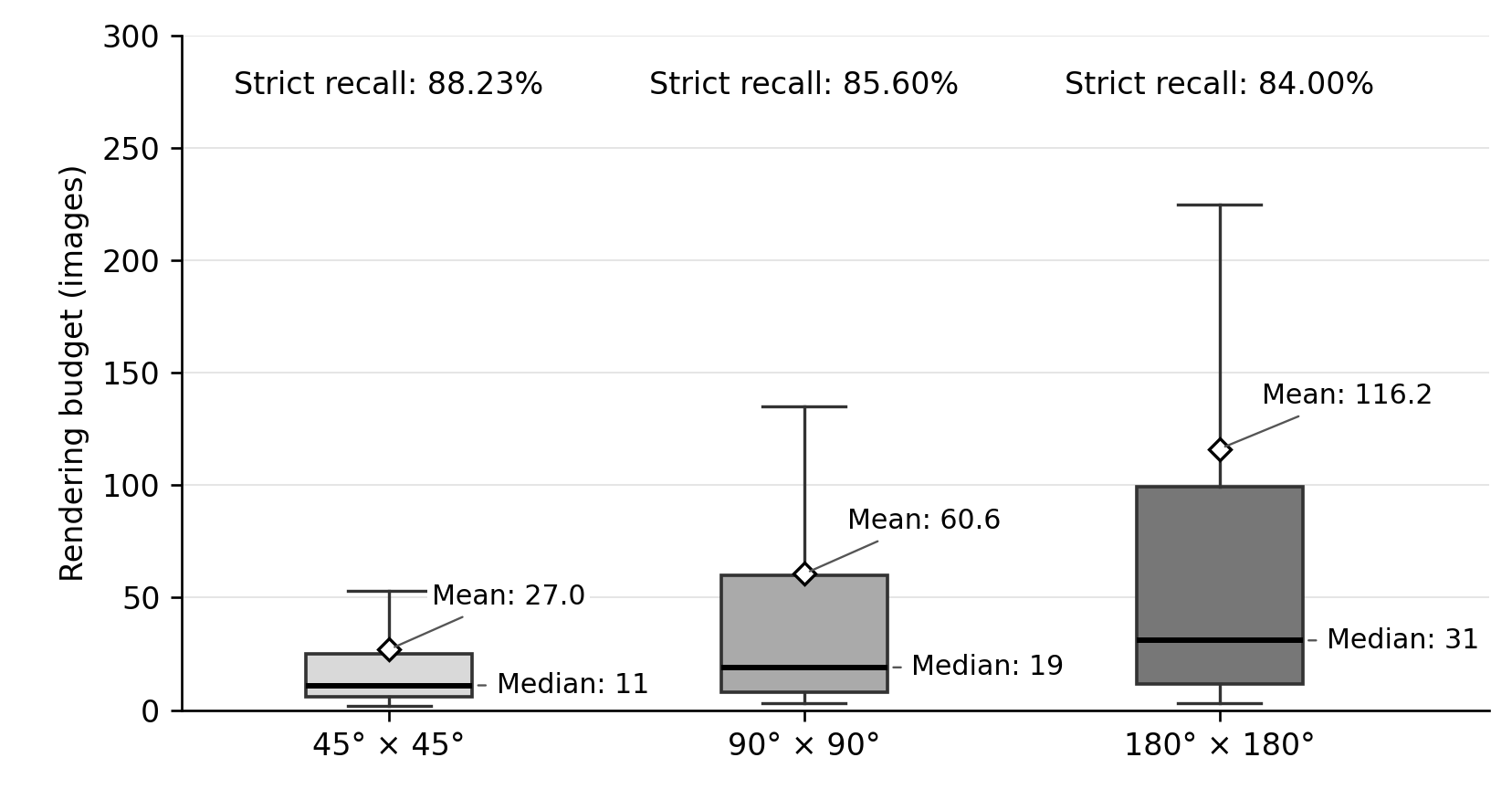}
\caption{\textbf{LunarGloc scalability across three search regions.}
Box plots show the number of additional online renders per query. Diamonds
mark means, horizontal lines mark medians, and percentages indicate strict
recall.}\label{fig:larger-scale_loc}
\end{figure}

LunarGloc maintains 84.00\% strict recall when the search spans a lunar
hemisphere, compared with 88.23\% in the standard $45^\circ \times 45^\circ$
region (Fig.~\ref{fig:larger-scale_loc}). Across the
$45^\circ \times 45^\circ$, $90^\circ \times 90^\circ$, and
$180^\circ \times 180^\circ$ regions, the mean number of online renders per
query increases from 27.0 to 60.6 to 116.2. Even at the hemispheric scale,
half of the queries require at most 31 online renders. 

LunarGloc uses a single layer of reference views at 50\,km altitude to
initialize the search at every scale. For the $90^\circ \times 90^\circ$
region, this layer contains 3,603 views, compared with 9,789 and 16,553
views in the three- and five-layer static databases. For the hemispheric
search, LunarGloc uses 10,154 reference views and renders additional
viewpoints on demand. Replicating the same horizontal grid at three or five
altitudes would require 30,462 or 50,770 precomputed views, respectively.

\section{Qualitative comparison}
\label{sec:supp-qualitative}

\subsection{Interactive demo of LunarGloc}
We included a demo of LunarGloc: \texttt{lunargloc-demo.html}. It includes 10 examples: five from the $45^\circ \times 45^\circ$ search space, which includes a comparison of LunarGloc and LunarGloc-w/o-guided, highlighting the efficiency of the guided search. We also included five examples to show LunarGloc's searching process over the entire visible lunar hemisphere ($180^\circ \times 180^\circ$). 

Readers can step through each render to see where it was sampled, which matches triggered a guided proposal, and how the consensus set size changed. The final frame shows the SIFT correspondences retained by PnP and the position and rotation errors of the estimated pose.

\subsection{CBN vs. LunarGloc}
As discussed in the main paper, crater-based navigation (CBN) struggles
in crater-sparse terrain and in polar regions where shadows obscure crater
rims. STELLA~\cite{McLeod2025STELLA} uses a Mask R-CNN detector trained
on synthetic images, while STELLA-LFM uses a Faster R-CNN detection head
fine-tuned on real lunar images~\cite{NASAIBM2026CraterDetection}.
STELLA-LFM generally detects more craters, which we attribute in part to
training on real imagery. 

\begin{figure}[!htbp]
  \centering
  \includegraphics[width=\linewidth]{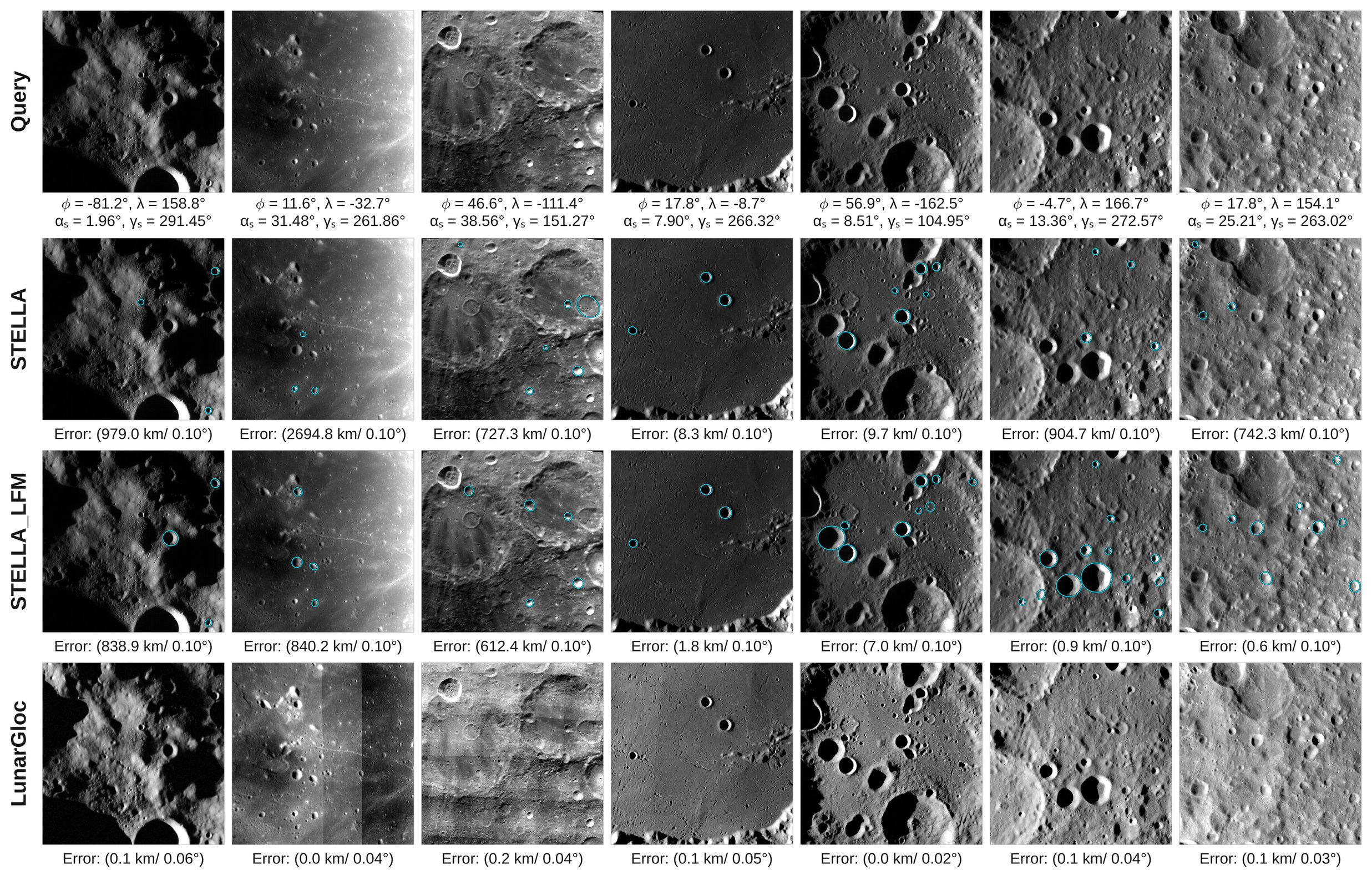}
  \caption{\textbf{Performance of CBN and LunarGloc on challenging query images.} Each column shows a query, the crater matches retained by STELLA and STELLA-LFM (cyan ellipses), and the view rendered at the final LunarGloc global localization module. With few crater matches, CBN's position errors can reach hundreds of kilometers. With the aid of other visual cues, such as other terrain features and shadows, LunarGloc manages to localize these challenging query images.}
  \label{fig:supp_STELLA_vs_LunarGloc}
\end{figure}

When too few detections have a counterpart in the crater catalog, the crater identification (CID) module of CBN can match them to the wrong set of cataloged craters, as seen in columns~1--3 of Fig.~\ref{fig:supp_STELLA_vs_LunarGloc}. Three factors can limit the number of matchable craters. Regions such as the lunar maria contain few craters, the mission catalog may omit some detected craters, and the detector may miss others (false negatives).

In the first two columns, both STELLA variants retain only three or four crater matches after CID, and their position errors exceed 800\,km, demonstrating the case where the detected craters are matched to the wrong cataloged craters. The last two columns show how additional detections can help. STELLA retains only 4 and 3 matches, respectively, with position errors exceeding 700\,km in both cases. STELLA-LFM retains 13 and 9 matches, respectively, and localizes both queries to within 1\,km. LunarGloc localizes accurately even in the examples with few crater matches, where other terrain features and shadows provide visual cues for pose estimation.

\clearpage

\subsection{UAVLoc vs. LunarGloc-Static}
\begin{figure}[!htbp]
  \centering
  \includegraphics[width=\linewidth]{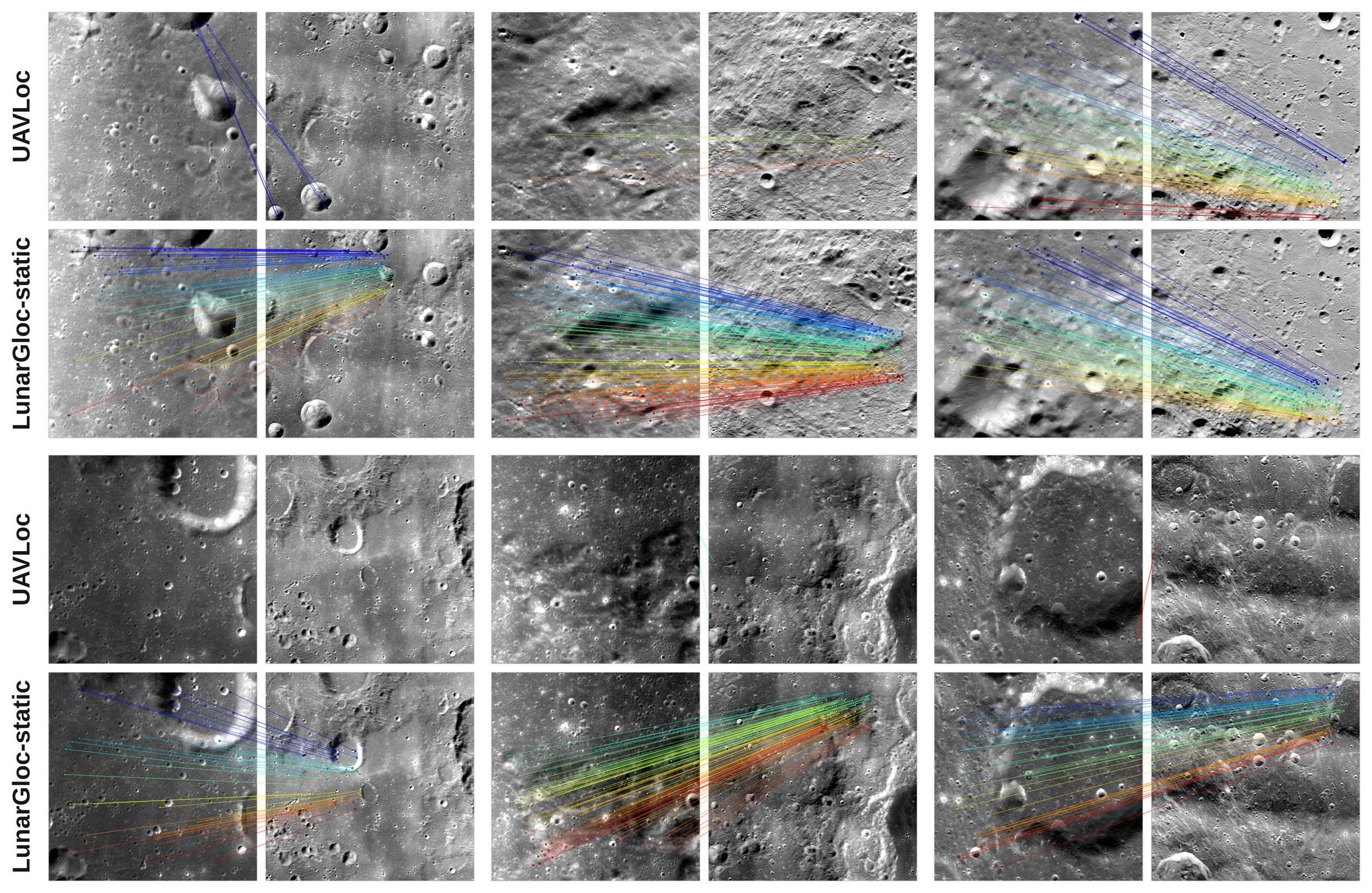}
  \caption{\textbf{Local feature matching on the same retrieved references.}
Each panel pairs a query (left) with a retrieved reference (right).
Within each pair of rows, lines show PnP inliers for UAVLoc (upper row)
and LunarGloc-Static (lower row). SIFT yields more inliers than
SuperPoint--SuperGlue on all six pairs shown.}
  \label{fig:uav_lgs}
\end{figure}

In our evaluation, UAVLoc and LunarGloc-Static use the same visual place
recognition module and retrieve the same top three references using the
combined score $s$. For pose refinement, UAVLoc matches
SuperPoint~\cite{DeTone2018SuperPoint} features with
SuperGlue~\cite{Sarlin2020SuperGlue}, while LunarGloc-Static uses
SIFT~\cite{Lowe2004SIFT}. Fig.~\ref{fig:uav_lgs} compares their matches
on six identical query-reference pairs. The difference is most apparent
in the lower half of the figure. From left to right, UAVLoc retains 0, 2,
and 2 PnP inliers on the displayed pairs, compared with 38, 78, and 40 for
LunarGloc-Static. We attribute UAVLoc’s lower recall (Tab.~\ref{tab:localization-comparison} of the main paper) to the smaller number of correspondences it obtains from the same retrieved references.

\clearpage

\subsection{MCLoc vs. LunarGloc}
\begin{figure}[!htbp]
  \centering
  \includegraphics[width=\linewidth]{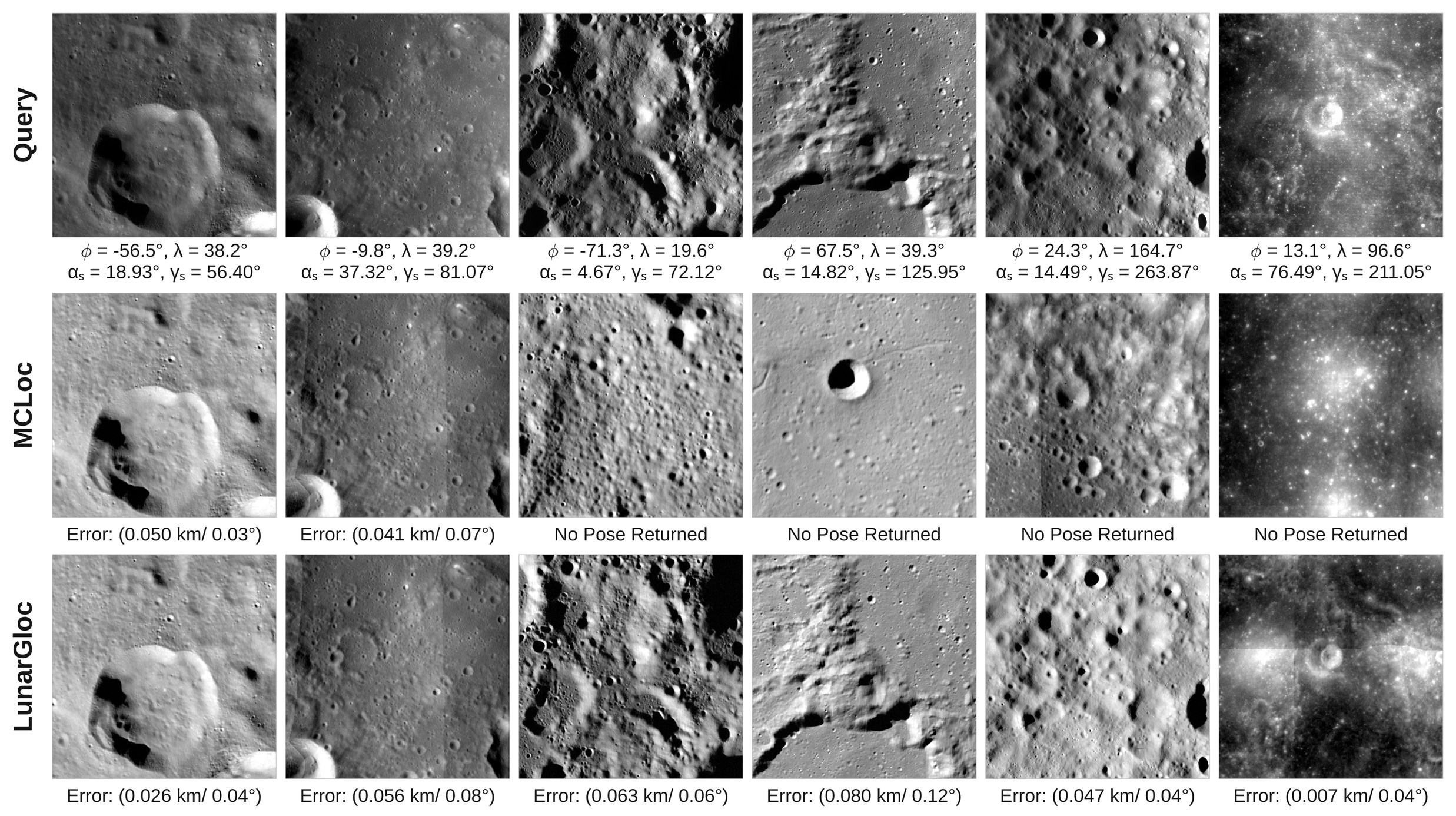}
\caption{\textbf{Candidate views selected before pose refinement.}
Each column shows a query (top) and the views selected by MCLoc (middle)
and LunarGloc (bottom). Annotations report position and attitude errors
after refinement. Both methods localize the first two queries. For the
remaining four, only LunarGloc returns a pose.}
  \label{fig:MCLoc_vs_LunarGloc}
\end{figure}

MCLoc tends to converge to the wrong location before the final pose refinement module. These are illustrated in the third to sixth column in Fig.~\ref{fig:MCLoc_vs_LunarGloc}. As observed, the converged location possess similar structure and illumination conditions as the query images. The final pose refinement module rejected these candidates since they have insufficient inliers. 

\subsection{LunarGloc's failure cases}
\begin{figure}[!htbp]
  \centering
  \includegraphics[width=\linewidth]{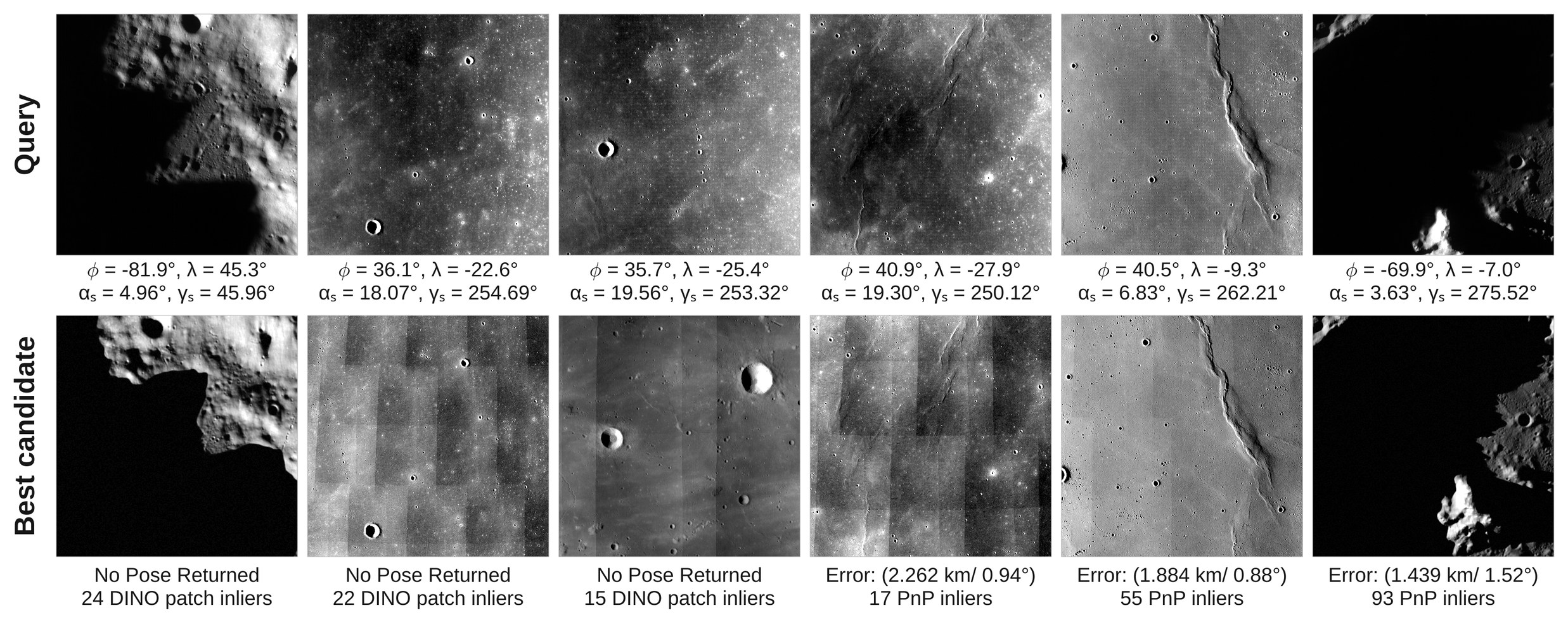}
\caption{\textbf{Failure cases of LunarGloc.} Each column pairs a query (top) with a rendered candidate (bottom). In the first three columns, the best candidates found during the search yield 24, 22, and 15 patch inliers, respectively. The last three columns show the accepted candidates, annotated with the final position and attitude errors and the number of PnP inliers.}
  \label{fig:lg_failure_cases}
\end{figure}

Fig.~\ref{fig:lg_failure_cases} illustrates two failure modes of LunarGloc. In the first three columns, LunarGloc exhausts its online rendering budget without returning a pose. The best candidate for each query falls short of the 25 patch inliers required to invoke pose refinement (Sec.~\ref{sec:supp-hyperparameters}). In the last three columns, LunarGloc returns a pose, but the final position error exceeds 1\,km in each case. These queries contain few distinctive terrain features, or the visible features are concentrated in a small part of the image.

\end{document}